\documentclass[final]{cvpr}

\usepackage{times}
\usepackage{epsfig}
\usepackage{graphicx}
\usepackage{amsmath}
\usepackage{amssymb}
\usepackage{adjustbox}

\usepackage{kotex}
\usepackage{rotating}
\usepackage[symbol]{footmisc}

\usepackage[pagebackref=true,breaklinks=true,colorlinks,bookmarks=false]{hyperref}
\pdfoutput=1

\begin{document}

\title{DRANet: Disentangling Representation and Adaptation Networks \\ for Unsupervised Cross-Domain Adaptation}

\author{Seunghun Lee\\
DGIST\\
{\tt\small lsh5688@dgist.ac.kr}
\and
Sunghyun Cho\\
POSTECH CSE \& GSAI\\
{\tt\small s.cho@postech.ac.kr}
\and
Sunghoon Im\thanks{Corresponding author.}\\
DGIST\\
{\tt\small sunghoonim@dgist.ac.kr}
}

\maketitle

\begin{abstract}
     In this paper, we present DRANet, a network architecture that disentangles image representations and transfers the visual attributes in a latent space for unsupervised cross-domain adaptation. Unlike the existing domain adaptation methods that learn associated features sharing a domain, DRANet preserves the distinctiveness of each domain's characteristics. Our model encodes individual representations of content (scene structure) and style (artistic appearance) from both source and target images. Then, it adapts the domain by incorporating the transferred style factor into the content factor along with learnable weights specified for each domain. This learning framework allows bi-/multi-directional domain adaptation with a single encoder-decoder network and aligns their domain shift. Additionally, we propose a content-adaptive domain transfer module that helps retain scene structure while transferring style. Extensive experiments show our model successfully separates content-style factors and synthesizes visually pleasing domain-transferred images. The proposed method demonstrates state-of-the-art performance on standard digit classification tasks as well as semantic segmentation tasks.

\end{abstract}

\vspace{-3mm}
\section{Introduction}
\label{sec:intro}

\begin{figure}[t] 
	\centering
    \includegraphics[width=0.84\linewidth]{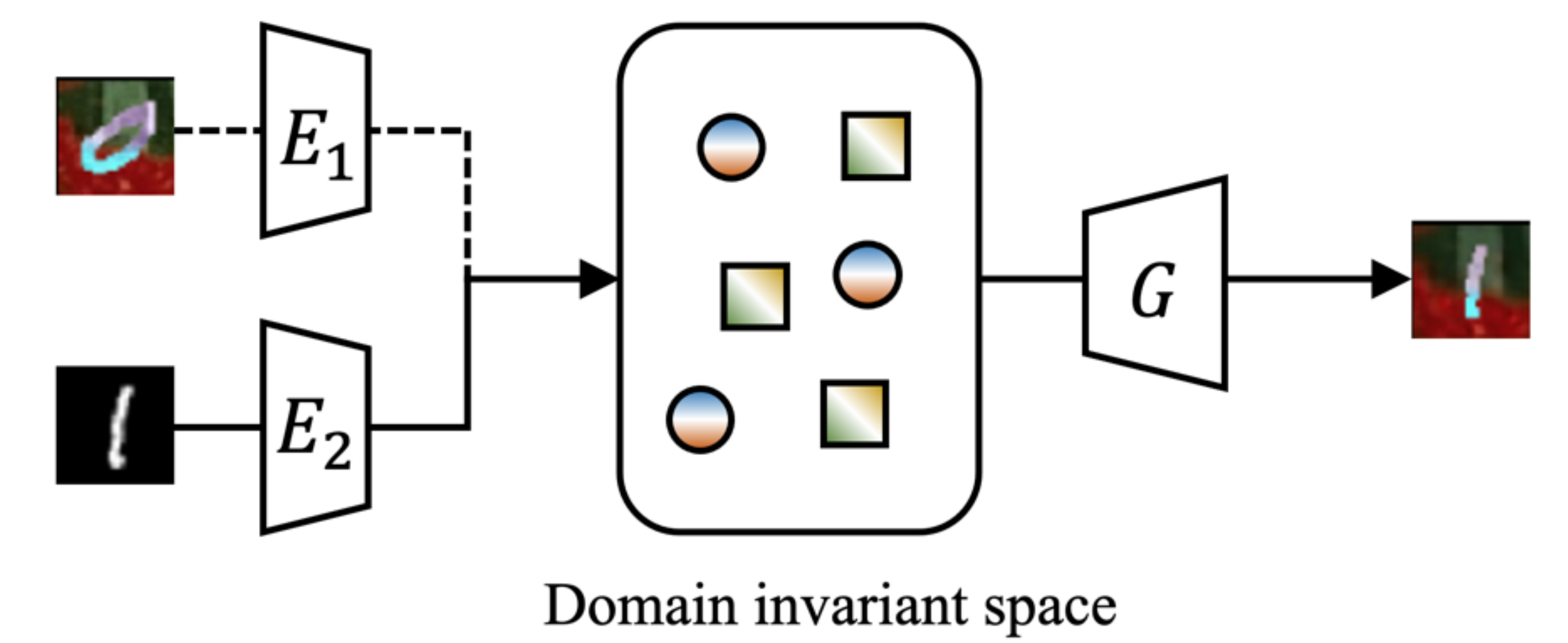}\\
    \small{(a) Traditional domain adaptation~\cite{ganin2016domain,hoffman2018cycada}} \\
    \includegraphics[width=0.84\linewidth]{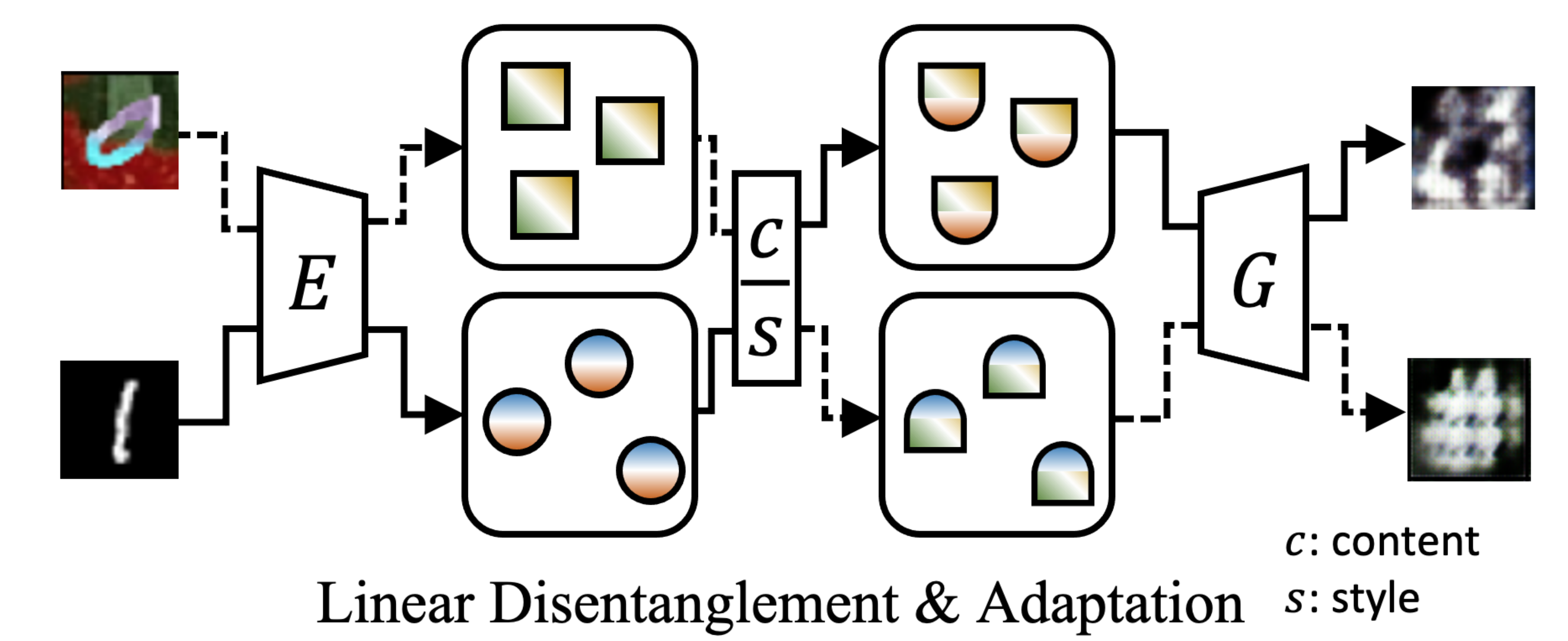}\\
    \small{(b) Linear feature separation~\cite{zhang2018style}} and domain adaptation\\
    \includegraphics[width=0.84\linewidth]{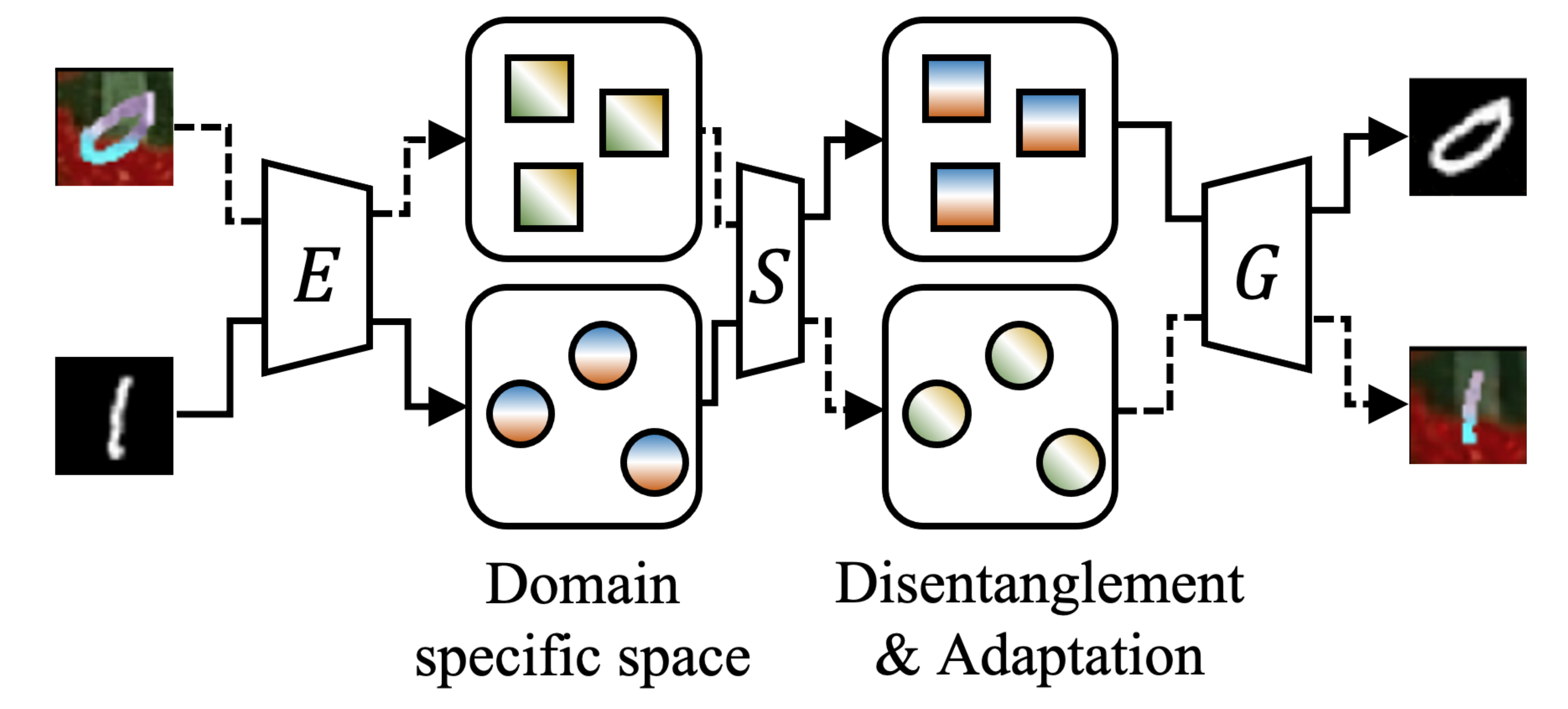}\\
    \small{(c) Our feature separation and domain adaptation (\textit{DRANet})}\\
    \vspace{2mm}
	\caption{Illustration of \textit{DRANet} and the competitive methods (domain adaptation~\cite{ganin2016domain,hoffman2018cycada}, representation disentanglement~\cite{zhang2018style}. Note that $E$, $S$, and $G$ are an encoder, a separator, and a generator.
	 }
	\label{fig:teaser}
	\vspace{-3mm}
\end{figure}

The use of deep neural networks (DNN) has led to significant performance improvements in a variety of areas, including computer vision~\cite{deng2009imagenet}, machine learning~\cite{goodfellow2016deep}, and natural language processing~\cite{devlin2018bert}.
However, problems remain, particularly domain gaps between data, which can significantly degrade model performance.
Extensive efforts have been made to generalize the models across domains using unsupervised domain adaptation~\cite{ben2006analysis,tzeng2014deep,long2015learning,tzeng2015simultaneous,ganin2016domain,sun2016deep,tzeng2017adversarial,liu2016coupled,bousmalis2017unsupervised,hoffman2018cycada,ye2020light}.
Unsupervised domain adaptation attempts to align the distribution shift in labeled source data with unlabeled target data.
Various strategies have been explored to bridge the gap across domains, for example, by feature learning and generative pixel-level adaptation.

Feature-level methods~\cite{tzeng2014deep,long2015learning,sun2016deep,tzeng2015simultaneous,ganin2016domain,sun2016deep,tzeng2017adversarial} learn features that combine task-discrimination and domain-invariance, where both domains are mapped into a common feature space.
Domain invariance typically involves minimizing some feature distance metric~\cite{tzeng2014deep,long2015learning,sun2016deep} or adversarial discriminator accuracy~\cite{ganin2016domain}.
Pixel-level approaches~\cite{liu2016coupled,bousmalis2017unsupervised} perform a similar distribution alignment, not in a feature space but in the raw pixel space by leveraging the power of Generative Adversarial Networks (GANs)~\cite{goodfellow2014generative, mirza2014conditional, radford2015unsupervised, salimans2016improved, chen2016infogan}.
They adapt source domain images so that they appear as if drawn from the target domain.
Some studies~\cite{hoffman2018cycada,tran2019gotta,ye2020light} incorporate both pixel-level and feature-level approaches to achieve complementary benefits.

Recently, the field of study has been further advanced by learning disentangled representations into the exclusive and shared components in a latent feature space~\cite{bousmalis2016domain,gonzalez2018image,liu2018detach,zou2020joint}.
They demonstrate that representation disentanglement improves a model's ability to extract domain invariant features, as well as the domain adaptation performance.
However, these methods still focus on the associated features between two domains such as shared and exclusive components, so they require multiple encoders and generators specialized in individual domains.
Moreover, the network training relies heavily on a task classifier with ground-truth class labels, in addition to domain classifiers.

To tackle these issues, we propose DRANet, a single feed-forward network, that does not require any ground-truth task labels for cross-domain adaptation.
In contrast to previous approaches in~\figref{fig:teaser}-(a) that map all domain images into a shared feature space, we focus on extracting the domain-specific features that preserve individual domain characteristics in~\figref{fig:teaser}-(c).
Then, we disentangle the discriminative features of individual domains into the content and style components using a separator, which are later used to generate the domain-adaptive features.
Unlike the previous feature separation work~\cite{zhang2018style}, which linearly divides latent vectors into two components in~\figref{fig:teaser}-(b), our separator is tailored to disentangle latent variables in a nonlinear manifold.
Our intuition behind the network design is that different domains may have different distributions for their contents and styles, which cannot be effectively handled by the linear separation of latent vectors.
Thus, to handle such difference, our network adopts the non-linear separation and domain-specific scale parameters that are dedicated to handle such inter-domain difference.

To the best of our knowledge, DRANet is the first approach based solely on the individual domain characteristics for unsupervised cross-domain adaptation.
It enables us to apply a single encoder-decoder network for a multi-directional domain transfer from fully unlabeled data.
The distinctive points of our approach are summarized as follows:
\vspace{-1mm}
\begin{itemize}
\setlength\itemsep{0em}
\item~We present DRANet, which disentangles image representation and adapts the visual attributes in a latent space to align the domain shift. 
\vspace{-1mm}
\item~We propose a content-adaptive domain transfer module that helps to synthesize realistic images of complex segmentation datasets, such as CityScapes~\cite{cordts2016cityscapes} and GTA5~\cite{richter2016playing}.
\vspace{-1mm}
\item~We demonstrate that images synthesized by our approach boost the task performances and achieve state-of-the-art performance on standard digit classification tasks as well as semantic segmentation tasks.
\end{itemize}

\section{Related Work}
\label{sec:related}

\subsection{Unsupervised Domain Adaptation}
Feature-level domain adaptation methods typically align learning distribution by modifying the discriminative representation space.
The strategy is to guide feature learning by minimizing the difference between the feature space statistics of the source and target.
Early deep adaptive approaches minimize some measurements of domain shift such as maximum mean discrepancy~\cite{tzeng2014deep,long2015learning} or correlation distances~\cite{sun2016deep}. 
Recent works~\cite{ganin2016domain,tzeng2015simultaneous,tzeng2017adversarial} learn the representation that is discriminative of source labels while not being able to distinguish between domain using an adversarial loss inspired by the work \cite{ben2006analysis}.
The domain-invariant features are discovered using standard backpropagation training with minimax loss~\cite{ganin2016domain}, domain confusion loss~\cite{tzeng2015simultaneous}, or GAN loss~\cite{tzeng2017adversarial}.

Another approach to unsupervised domain adaptation is the generative pixel-level domain adaptation, which synthesizes images with the content of source images and the style of target images using the adversarial training~\cite{goodfellow2014generative}.
Liu and Tuzel~\cite{liu2016coupled} accomplish to learn the joint distribution of source and target representations by weight sharing, using a specific layer responsible for decoding abstract semantics.
Bousmalis \etal~\cite{bousmalis2017unsupervised} use GANs to learn transformations in the pixel space from one domain to another. 
Hoffman \etal~\cite{hoffman2018cycada} adapt representations both at the pixel and feature levels while enforcing both structural and semantic consistency using a cycle-consistency loss.
Ye \etal~\cite{ye2020light} also incorporate both pixel and feature-level domain classifiers to calibrate target domain images whose representations are close to those of the source domain.

\begin{figure*}[t] 
	\centering
    \includegraphics[width=0.95\linewidth]{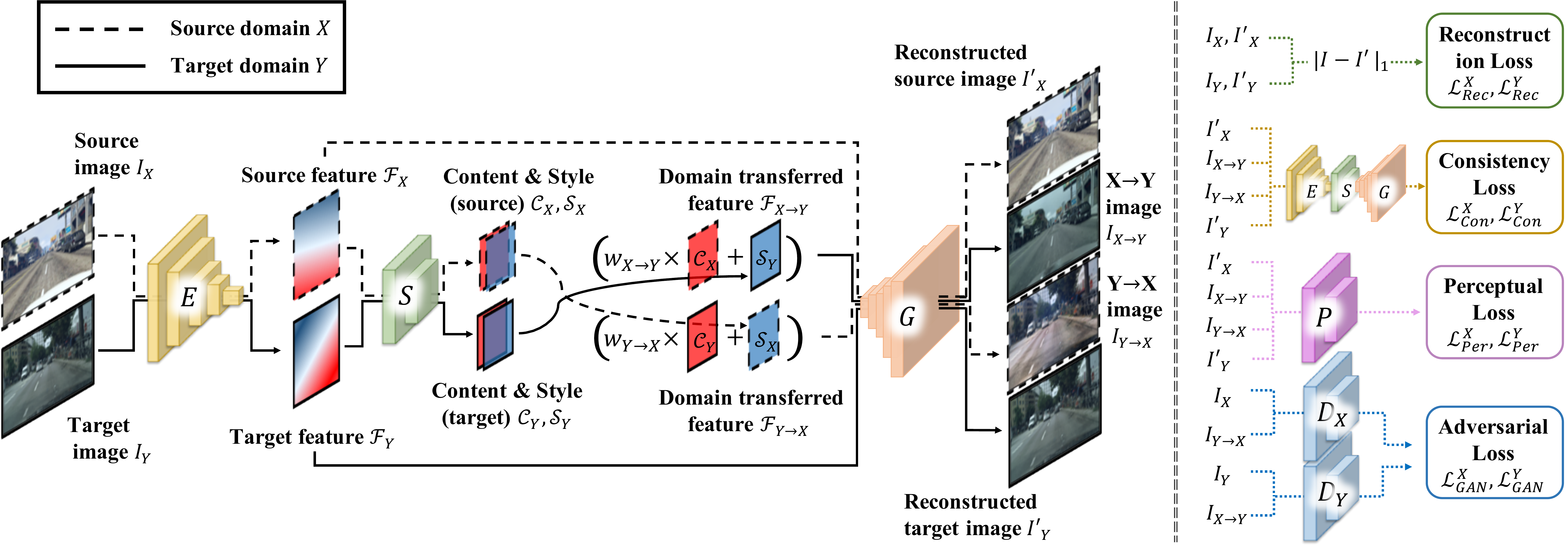}
	\caption{Overview of our model. (Left) Image translation blocks involving an encoder $E$, a separator $S$, and a generator $G$. The source and target images $I_X, I_Y$ are the input, and the reconstructed images $I'_X, I'_Y$ and domain transferred images $I_{X\to Y}, I_{Y\to X}$ are the output. (Right) The training losses involving reconstruction $\mathcal{L}_{Rec}$, consistency $\mathcal{L}_{Con}$, perceptual $\mathcal{L}_{Per}$, and adversarial $\mathcal{L}_{GAN}$ loss.}
	\label{fig:overview}
	\vspace{-3mm}
\end{figure*}

\subsection{Disentangling Internal Representation}
The separation of style and content components in a latent space has been widely studied for artistic style transfer~\cite{tenenbaum2000separating,elgammal2004separating,gatys2016image,zhang2018style,zhang2018separating}.
Tenebaum and Freeman~\cite{tenenbaum2000separating} show how perceptual systems can separate the content and style factors, and propose bilinear models to solve these two factors.
Elgammal and Lee~\cite{elgammal2004separating} introduce a method to separate style and content on manifolds representing dynamic objects.
Gatys \etal~\cite{gatys2016image} show how generic feature representations learned by a CNN manipulate the content and style of natural images.
Zhang \etal~\cite{zhang2018separating} propose a neural network representing each style and content with a small set of images, while separating the representations.
Zhang \etal~\cite{zhang2018style} bimodally divide feature representations into the content and style components. 

Among the studies on domain adaptation, the search for approaches to disentangling internal representations has recently grown in interest.
Bousmalis \etal~\cite{bousmalis2016domain} learn to extract image representations that are partitioned into two subspaces: private and shared components and show that the modeling of unique features helps to extract domain-invariant features.
Gonzalez-Garcia \etal~\cite{gonzalez2018image} attempt to disentangle factors that are exclusive in both domains, and factors that are shared across domains.
Liu \etal~\cite{liu2018detach} propose a cross-domain representation disentangler that bridges the information across data domains and transfers the attributes.
Zou \etal~\cite{zou2020joint} introduce a joint learning framework that separates id-related/unrelated features for person re-identification tasks.
We discuss the major differences between our work and the listed works in~\secref{sec:intro}.

\section{DRANet}
\label{sec:method}

\subsection{Overview}
\label{sec:overview}
\begin{figure}[t] 
	\centering
	\begin{tabular}{c@{\hspace{1mm}}}
    \includegraphics[width=0.7\linewidth]{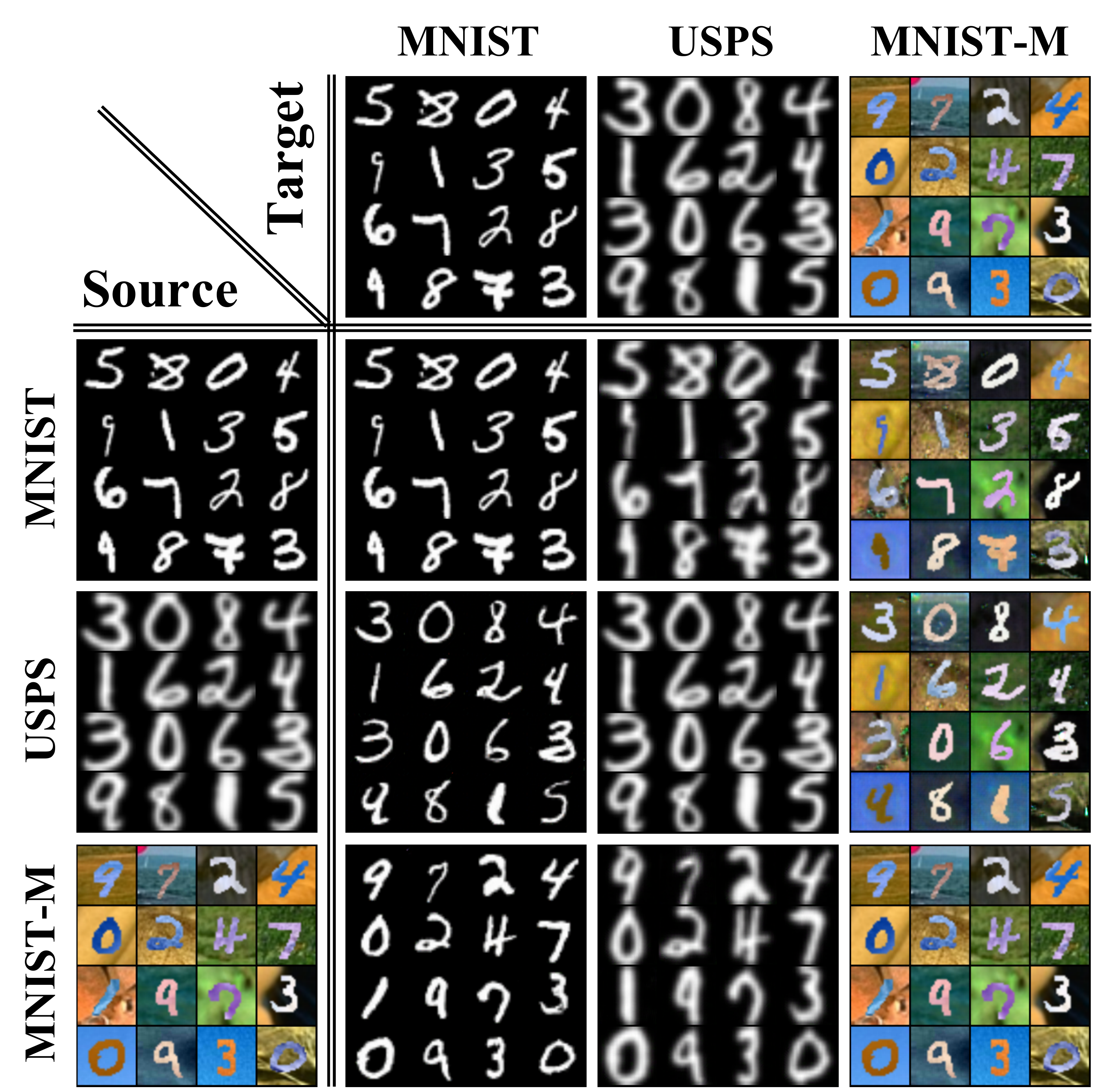}
	\end{tabular}
	\caption{Tri-directional domain adaptation results from our single network. DRANet keeps the content of source images and transfers the domain of target images.}
	\label{fig:3domain}
	\vspace{-3mm}
\end{figure}

The overall pipeline of our method is illustrated in~\figref{fig:overview}.
Our framework can be extended to domain transfer across three domains, as shown in~\figref{fig:3domain}, although the example only shows two domain case for simple illustration. 
The networks consist of an encoder $E$, a feature separator $S$, a generator $G$, two discriminators of the source and target domains $D_X, D_Y$, and a perceptual network $P$.
In the training phase, we learn all of the parameters of these networks, as well as the feature scaling factors $w_{X \to Y}, w_{Y \to X}$ which compensate for the distribution of two domains.
Given the source and target images $I_X, I_Y$, the encoder $E$ extracts the individual features $\mathcal{F}_X, \mathcal{F}_Y$ that later pass through the generator $G$ to reconstruct the original input images $I'_X, I'_Y$.
The separator $S$ disentangles each feature $\mathcal{F}_X, \mathcal{F}_Y$ into the components of scene structure and artistic appearance, which in this paper we call the content $\mathcal{C}_X, \mathcal{C}_Y$ and the style $\mathcal{S}_X, \mathcal{S}_Y$, respectively.
Then, the transferred domain features $\mathcal{F}_{X \to Y}, \mathcal{F}_{Y \to X}$ are synthesized with the learnable scale parameters $w_{X \to Y}, w_{Y \to X}$.
The generator $G$ maps the original features $\mathcal{F}_X, \mathcal{F}_Y$ and the transferred features $\mathcal{F}_{X \to Y}, \mathcal{F}_{Y \to X}$ into their image space $I'_X, I'_Y, I_{X \to Y}, I_{Y \to X}$, respectively.
The pretrained perceptual network $P$, extracts perceptual features to impose the constraints on both content similarity and style similarity.
We use two discriminators, $D_X, D_Y$, to impose the adversarial loss on both domains.
In the test phase, just the encoder $E$, the separator $S$, the generator $G$, and domain weights $w$ are used to produce domain transferred images $I_{X\to Y}, I_{Y\to X}$ given source and target images $I_X, I_Y$.
With the single feed-forward network $E$-$S$-$G$, our method enables the bi-directional domain transfer of input images. 

\subsection{Disentangling Representation and Adaptation}
\label{sec:Network}
In this subsection, we describe the motivation for the design of our separator $S$.
We first extract the individual image features $\mathcal{F}_X,\mathcal{F}_Y$ using the weight-shared encoder:
\begin{gather}
\begin{split}
\mathcal{F}_X = E(I_X),~\mathcal{F}_Y = E(I_Y).
\end{split}
\end{gather}
The separator disentangles these features into scene structure and artistic appearance factors.
We hypothesize that the nonlinear manifold learning is still necessary to map each domain-specific representation into the content or style spaces as demonstrated in~\cite{elgammal2004separating}.
Thus, we learn a non-linear projection function $S$ that separates the features $\mathcal{F}_X$ into content $\mathcal{C}_X$ and style $\mathcal{S}_X$ factors, as follows:
\begin{gather}
\begin{split}
\mathcal{C}_X = w_XS(\mathcal{F}_X),~
\mathcal{S}_X = \mathcal{F}_X-w_XS(\mathcal{F}_X),
\end{split}
\end{gather}
where $w_X$ is the weight parameter that normalizes the distribution of content space, which helps to compensate for the distribution shift.
The content component is obtained using the non-linear function and the learnable feature scaling parameters, while the style component is defined by subtracting content components from the whole feature.
The target representation $\mathcal{F}_Y$ is also passed through the same separator $S$, and outputs the target content and style $\mathcal{C}_Y, \mathcal{S}_Y$, but for simplicity here we only denote the source domain case.

The disentangled representation is used to transfer the domain of features across domains as follows:
\begin{gather}
\begin{split}
\label{eq:feature}
\mathcal{F}_{X \to Y} = w_{X \to Y} \mathcal{C}_X + \mathcal{S}_Y, ~~
\mathcal{F}_{Y \to X} = w_{Y \to X} \mathcal{C}_Y + \mathcal{S}_X, \\
\text{where}~w_{X \to Y} = \frac{w_Y}{w_X},~w_{Y \to X} = \frac{w_X}{w_Y}.
\end{split}
\end{gather}
In our implementation, we directly learn the relative scale parameters $w_{X \to Y}, w_{Y \to X}$ along with all model parameters.
Finally, we pass all representations involving the domain adaptive features $\mathcal{F}_{X\to Y}, \mathcal{F}_{Y\to X}$ and the original source and target features $\mathcal{F}_X,\mathcal{F}_Y$ through the generator $G$ to project them into image space as follows:
\begin{gather}
\begin{split}
I_{X\to Y} = G(\mathcal{F}_{X\to Y}),~I_{Y\to X} = G(\mathcal{F}_{Y\to X}),\\
I'_X = G(\mathcal{F}_X),~I'_Y = G(\mathcal{F}_Y),
\end{split}
\end{gather}
where $I_{X\to Y}, I_{Y\to X}$ are the domain adapted images and $I'_X, I'_Y$ are the reconstructed images.

\subsection{Content-Adaptive Domain Transfer (CADT) }
Style transfer tends to struggle with the complex scenes containing various objects, such as a driving scene.
This is because those images are composed of different scene structure, as well as various object composition.
To tackle this problem, we present a Content-Adaptive Domain Transfer (CADT).
The key idea of this module is to search the target features whose content component is most similar to the source features.
Then, the domain transfer is conducted by reflecting more style information from more suitable target features.
To achieve this, we design a content similarity matrix for the database in a mini-batch, as follows:
\begin{gather}
\begin{split}
\label{eq:csim}
\mathcal{H}_{row} = \sigma _{row} \left( \mathbf{\mathcal{C}}_X \cdot \mathbf{\mathcal{C}}_Y^\top \right) = \left[ \begin{matrix} \mathcal{C}_{11} & \cdots & \mathcal{C}_{1b} \\ \vdots & \ddots & \vdots \\ \mathcal{C}_{b1} & \cdots & \mathcal{C}_{bb} \end{matrix} \right], \\~~
\mathbf{\mathcal{C}}_X, \mathbf{\mathcal{C}}_Y \in \mathbb{R}^{B \times N},
\end{split}
\end{gather}
where $\sigma_{row}$ is the softmax operation in the row dimension.
The size of the content factors $\mathcal{C}_X$ is defined by the batch size $B$ and the feature dimension $N$.
The matrix $\mathcal{H}_{row}$ contains information about the level of similarity between components in the mini-batch.
Based on the similarity matrix, we build a content-adaptive style feature as follows:
\begin{equation}
    \hat{\mathcal{S}}_Y = \mathcal{H}_{row}\mathcal{S}_Y, \text{where}~\mathcal{S}_Y \in \mathbb{R}^{B \times N}.
\end{equation}
More visually pleasing results can be expected than when using the normal transferring method because the content features are more likely to be stylized by the scenes containing similar structure and object composition.
We empirically demonstrate this in~\figref{fig:contentadaptive}.
To apply the content-adaptive domain transfer in the opposite direction, the content similarity matrix is simply obtained:
\begin{gather}
\begin{split}
\mathcal{H}_{col} = \left( \sigma_{col} \left( \mathbf{\mathcal{C}}_X \cdot \mathbf{\mathcal{C}}_Y^\top \right) \right)^\top,
\end{split}
\end{gather}
where $\sigma_{col}$ is the softmax in the column direction.

\subsection{Training Loss}
\label{sec:Loss}
We train our network with an encoder $E$, a separator $S$, and a generator $G$ by minimizing the loss function $\mathcal{L}^d$ while the discriminator $D_d$ tries to maximize it:
\begin{equation}
    \min_{E,S,G} \bigg(\sum_{d \in \{X, Y\}} \max_{D_d} \mathcal{L}^d\bigg),
\end{equation}
where the domain $d$ is either a source or target domain $X, Y$.
The overall loss of our framework consists of the reconstruction $\mathcal{L}_{Rec}$, consistency $\mathcal{L}_{Con}$, perceptual $\mathcal{L}_{Per}$, and adversarial $\mathcal{L}_{GAN}$ loss with the balancing term $\alpha_i$:
\begin{gather}
\begin{split}
    \mathcal{L}^d & = \alpha _1 \mathcal{L}_{Rec}^d + \alpha _2 \mathcal{L}_{GAN}^d + \alpha _3 \mathcal{L}_{Con}^d + \alpha _4 \mathcal{L}_{Per}^d.
\end{split}
\end{gather}
The followings are the details of each loss.

\textbf{Reconstruction Loss.} \quad
We impose an L1 loss to learn $E$ and $G$ that minimizes the difference between input image $I_d$ and the reconstructed image $I'_d$:
\begin{gather}
\begin{split}
    \mathcal{L}_{Rec}^d 
    = \mathcal{L}_1(I_d, I'_d),~\text{where}~I'_d=G(E(I_d)).
\end{split}
\end{gather}

\textbf{Adversarial Loss.} \quad
We apply two discriminators $D_{d\in \{X, Y\}}$ to evaluate the adversarial loss on the source and target domain, respectively.
The following is the adversarial loss for the domain adaptation of $X$ to $Y$:
\begin{gather}
\label{eq:Y}
\begin{split}
    \mathcal{L}_{GAN}^Y
    & = \mathbb{E}_{y \sim p_{data \left (Y \right ) }} \left[ \log D_Y(y) \right ] \\
    &+ \mathbb{E}_{(x,y) \sim p_{data \left (X, Y \right ) }} \left[ \log (1 - D_Y(I_{X \to Y}(x,y)) \right ]. \\
\end{split}
\end{gather}
We impose the same adversarial loss $\mathcal{L}_{GAN}^X$ for the adaptation of $Y$ to $X$ as well.
We apply spectral normalization \cite{miyato2018spectral} to all layers in G and D, and use PatchGAN Discriminator \cite{isola2017image} with the hinge version of adversarial loss \cite{lim2017geometric, tran2017deep,miyato2018cgans, zhang2019self} for driving scene adaptation.

\begin{figure}[t] 
	\centering
	\begin{tabular}{c@{\hspace{1mm}}c@{\hspace{1mm}}}
    \includegraphics[height=0.45\linewidth]{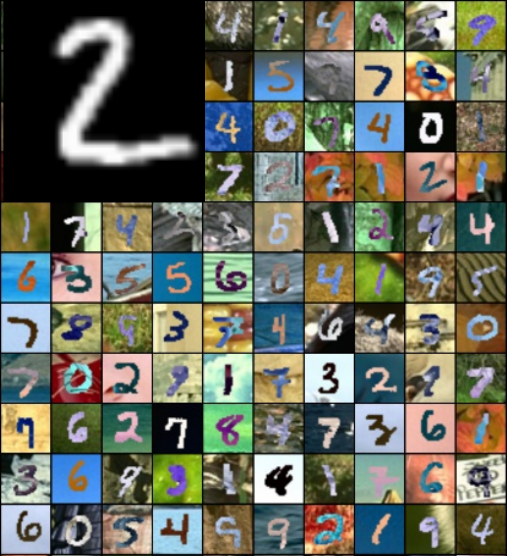} &     \includegraphics[height=0.45\linewidth]{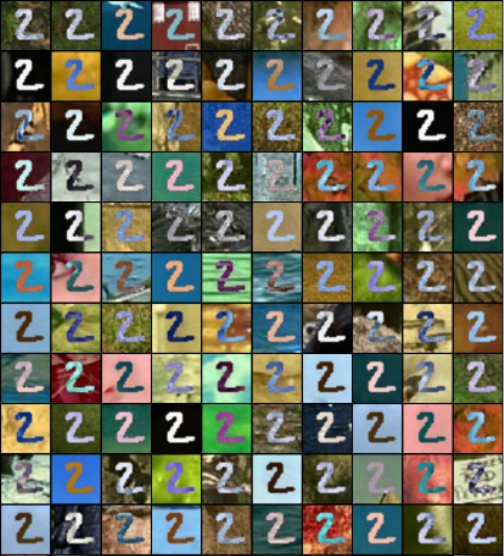} \\
    \small{(a) Source and target images} & \small{(b) Domain transferred images}
    \end{tabular}
	\caption{Various domain transferred examples from MNIST to MNIST-M. (a) Top-left image is source image of digit $2$ and the others are target images. (b) Domain transferred images.}
	\label{fig:transferred}
	\vspace{-3mm}
\end{figure}

\textbf{Consistency Loss.} \quad
The consistency loss attempts to retain the content and style components after re-projecting the domain transferred images into the representation space denoted as:
\begin{gather}
\begin{split}
    \mathcal{L}_{Con}^X = \mathcal{L}_1\big(\mathcal{C}_X, \mathcal{C}_{X\to Y}) + \mathcal{L}_1(\mathcal{S}_X, \mathcal{S}_{Y \to X}),\\ 
    \mathcal{L}_{Con}^Y = \mathcal{L}_1\big(\mathcal{C}_Y, \mathcal{C}_{Y\to X}) + \mathcal{L}_1(\mathcal{S}_Y, \mathcal{S}_{X \to Y}),
\end{split}
\end{gather}
where the content $C_{X\to Y}, C_{Y\to X}$ and style $S_{X\to Y}, S_{Y\to X}$ factors are extracted by passing the domain transferred images $I_{X\to Y}, I_{Y\to X}$, respectively, through the same encoder $E$ and separator $S$.
This loss explicitly encourages the scene structure consistency and artistic appearance consistency before and after domain adaptation.

\textbf{Perceptual Loss.} \quad
Conventionally, the GT class labels in (semi-)supervised training are provided as the semantic cues guiding the representation disentanglement.
However, our framework trains disentangling representations without any labeled data.
To learn the disentangler in an unsupervised manner, we impose a perceptual loss~\cite{johnson2016perceptual} which is widely known as a typical framework for style transfer, defined as:
\begin{gather}
\begin{split}
    \mathcal{L}_{Per}^X & = \mathcal{L}_{Content}^X + \lambda\mathcal{L}_{Style}^X, \\
    \mathcal{L}_{Per}^Y & = \mathcal{L}_{Content}^Y + \lambda\mathcal{L}_{Style}^Y, 
\end{split}
\end{gather}
where $\mathcal{L}_{Content}^X, \mathcal{L}_{Content}^Y$ are the content losses, and $\mathcal{L}_{Style}^X, \mathcal{L}_{Style}^Y$ are the style losses defined as:
\begin{gather}
\begin{split}
    \mathcal{L}_{Content}^Y & = \sum_{l \in L_C} \lVert P_l(I_X) - P_l(I_{X \to Y}) \rVert _2^2, \\
    \mathcal{L}_{Style}^Y & = \sum_{l \in L_S} \lVert \mathcal{G}(P_l(I_Y)) - \mathcal{G}(P_l(I_{X \to Y})) \rVert _F^2,
\end{split}
\end{gather}
where the set of layers $L_C, L_S$ are the subset of the perceptual network $P$. The weight parameter $\lambda$ balances the two losses, and $\mathcal{G}$ is the function that builds a Gram Matrix, given the features of each layer $l$~\cite{gatys2015texture}.
We also apply batch-instance normalization \cite{nam2018batch} for better stylization.
Details of architecture are described in the supplementary materials.

\begin{figure}[t] 
	\centering
	\begin{tabular}{c@{\hspace{1mm}}}
    \includegraphics[width=0.63\linewidth]{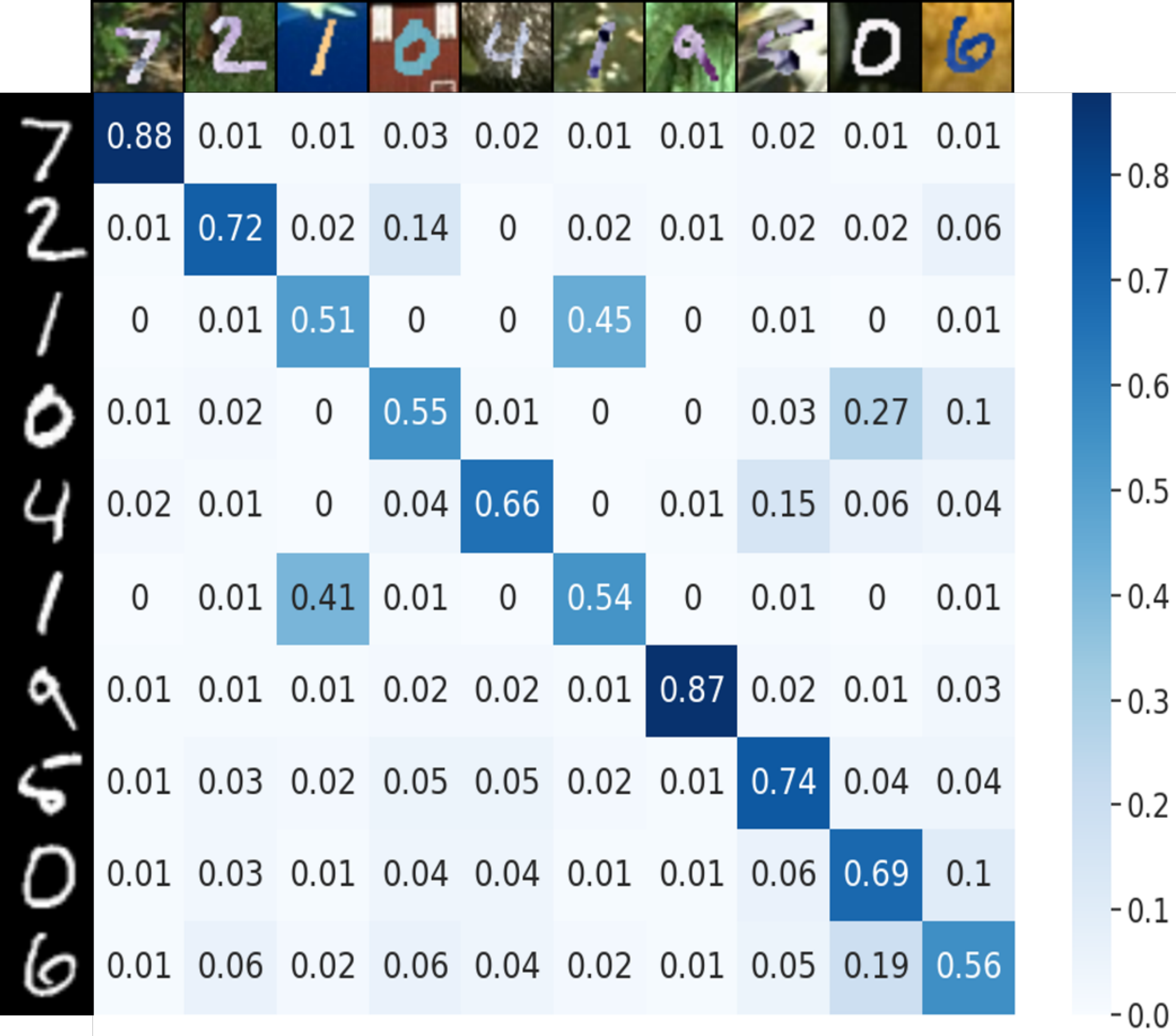} 
    \end{tabular}
	\caption{Content similarity between MNIST and MNIST-M.}
	\label{fig:confusion}
	\vspace{-3mm}
\end{figure}

\begin{table*}[t]
\begin{center}
\small
\begin{tabular}{lcccc}
\hline
\textbf{Method} & \textbf{MNIST to USPS} & \textbf{USPS to MNIST} & \textbf{MNIST to MNIST-M} & \textbf{MNIST-M to MNIST}  \\
\hline
Source Only~ & 80.2 & 44.9 & 62.5 & 97.8 \\
DANN~\cite{ganin2016domain} & 85.1 & 73.0 & 77.4 & -  \\
DSN~\cite{bousmalis2016domain} & 91.3 & - & 83.2 & -  \\
ADDA~\cite{tzeng2017adversarial} & 90.1 & 95.2 & - & -  \\
CoGAN~\cite{liu2016coupled} & 91.2 & 89.1 & 62.0 & -  \\
pixelDA~\cite{bousmalis2017unsupervised} & 95.9 & - & 98.2 & -  \\
CyCADA~\cite{hoffman2018cycada} & 95.6 & 96.5 & - & -  \\
LC + CycleGAN~\cite{ye2020light, zhu2017unpaired} & 97.1 & \textbf{98.3} & - & - \\
\hline
Ours (Bi-directional) & \textbf{98.2} & 97.8 & \textbf{98.7} & \textbf{99.3} \\
Ours (Tri-directional) & 97.6 & 96.9 & 98.3 & 99.0 \\
\hline
Target Only~ & 97.8 & 99.1 & 96.2 & 99.1 \\
\hline
\end{tabular}
\end{center}
\caption{Result comparison of DRANet to state-of-the-art methods on domain adaptation for digit classification. We report the performance from both bi-directional and tri-directional domain adaptation. Note that ours(bi-directional) and ours(tri-directional) use two models (MNIST-USPS, MNIST-MNISTM) and a model (MNIST-USPS-MNISTM), respectively to evaluate all four domain adaptation tasks.}
\label{tab:digit}
\end{table*}

\begin{figure*}[t] 
	\centering
	\begin{tabular}{c@{\hspace{1mm}}c@{\hspace{1mm}}c@{\hspace{1mm}}c@{\hspace{1mm}}}
    \includegraphics[width=0.2\linewidth]{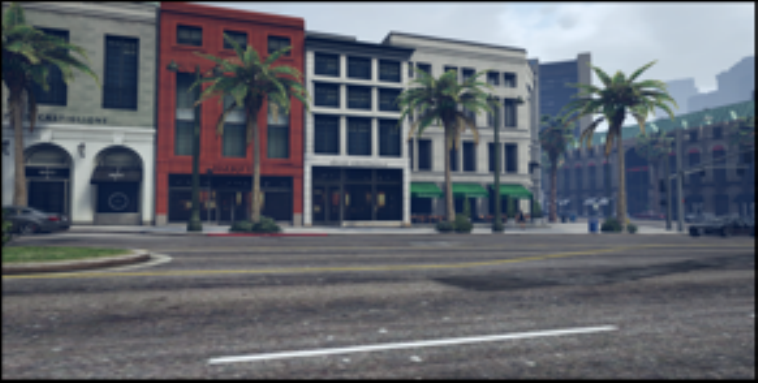}
    \includegraphics[width=0.2\linewidth]{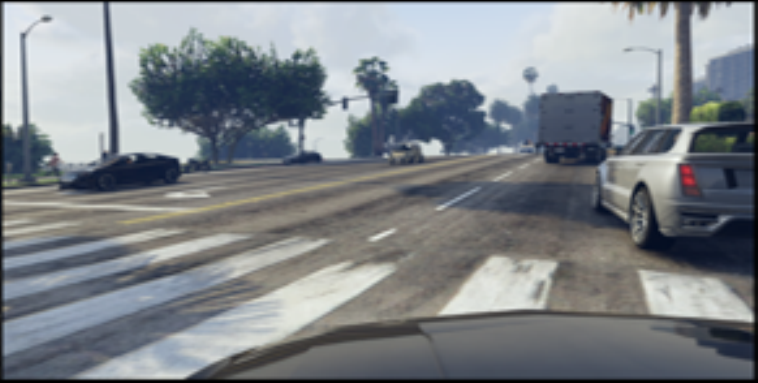}
    \includegraphics[width=0.2\linewidth]{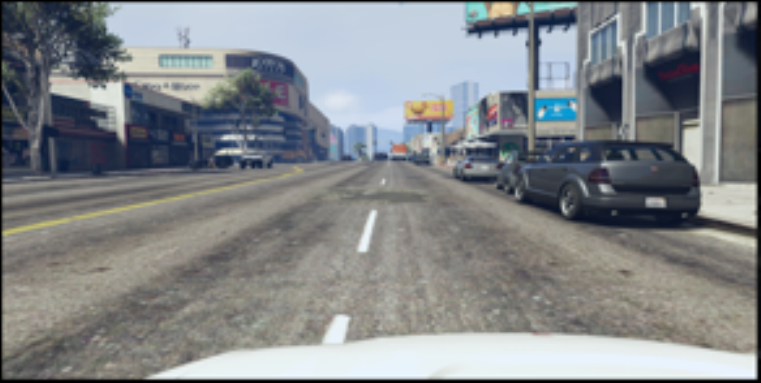}
    \includegraphics[width=0.2\linewidth]{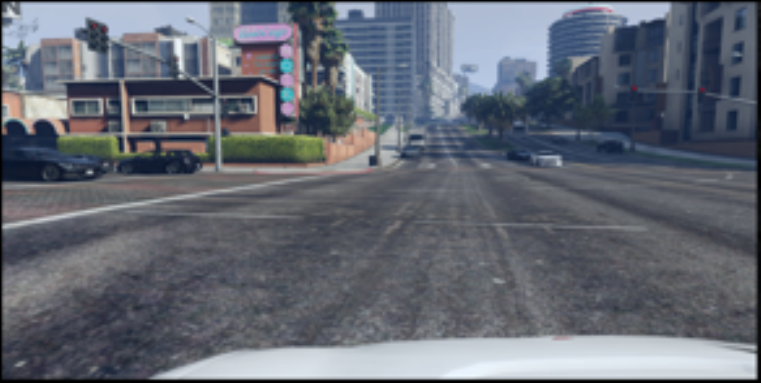} \\
    \small{(a) GTA5 original images}\\
    \includegraphics[width=0.2\linewidth]{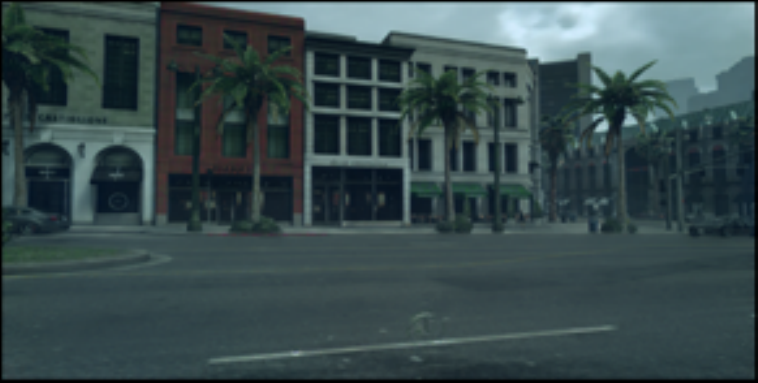}
    \includegraphics[width=0.2\linewidth]{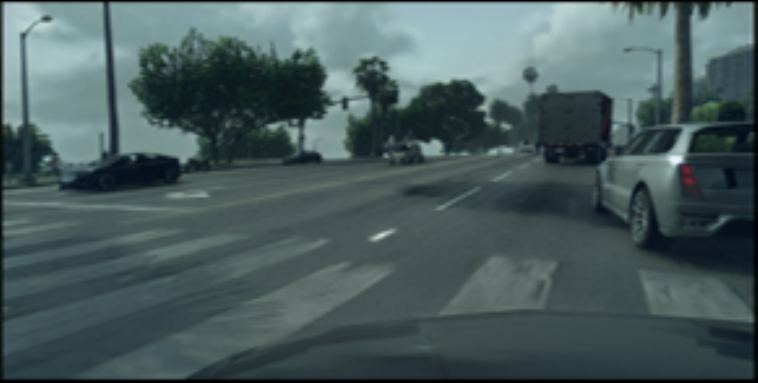}
    \includegraphics[width=0.2\linewidth]{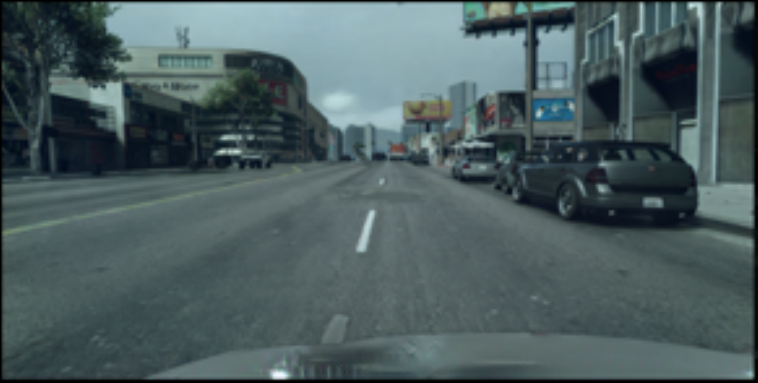}
    \includegraphics[width=0.2\linewidth]{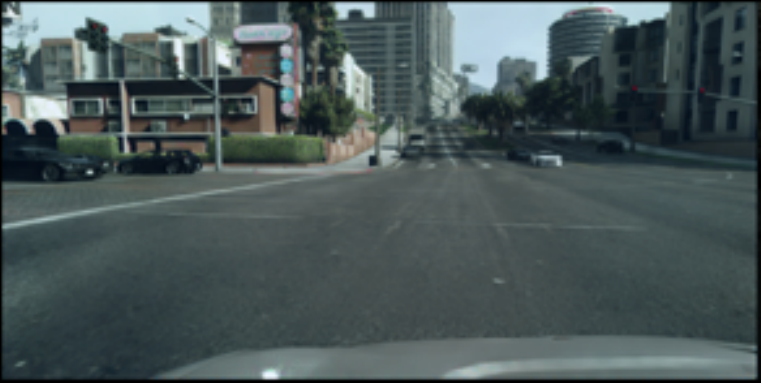} \\
    \small{(b) Transferred images using (a) GTA5 content and (c) CityScapes style.}\\
    \includegraphics[width=0.2\linewidth]{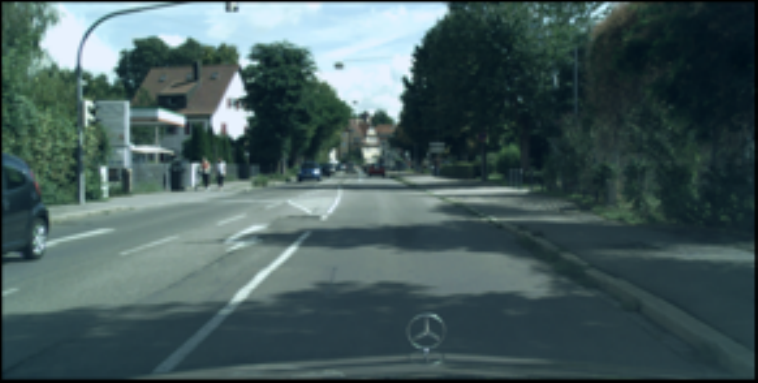}
    \includegraphics[width=0.2\linewidth]{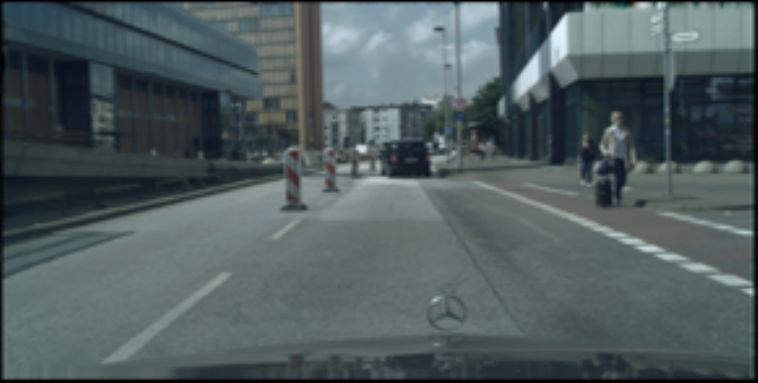}
    \includegraphics[width=0.2\linewidth]{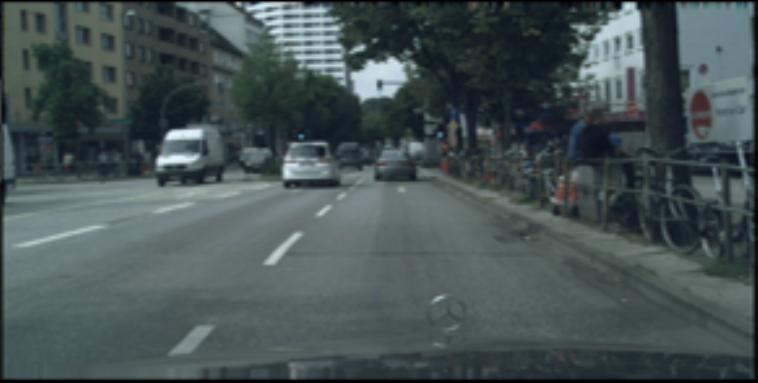}
    \includegraphics[width=0.2\linewidth]{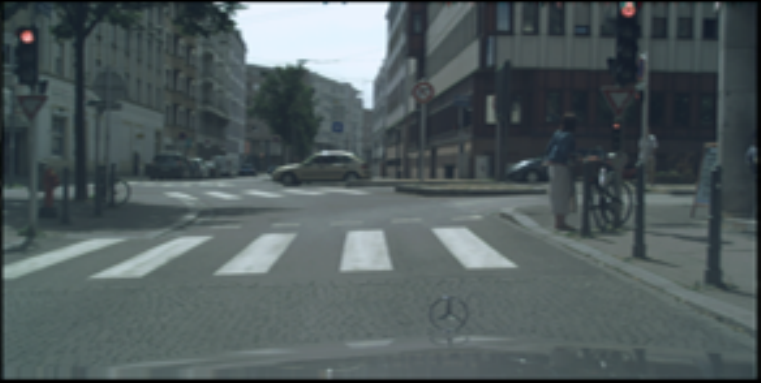}\\
    \small{(c) CityScapes original images}\\
    \includegraphics[width=0.2\linewidth]{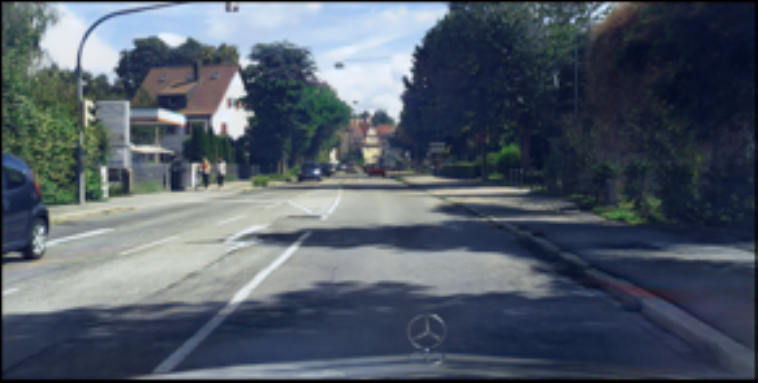}
    \includegraphics[width=0.2\linewidth]{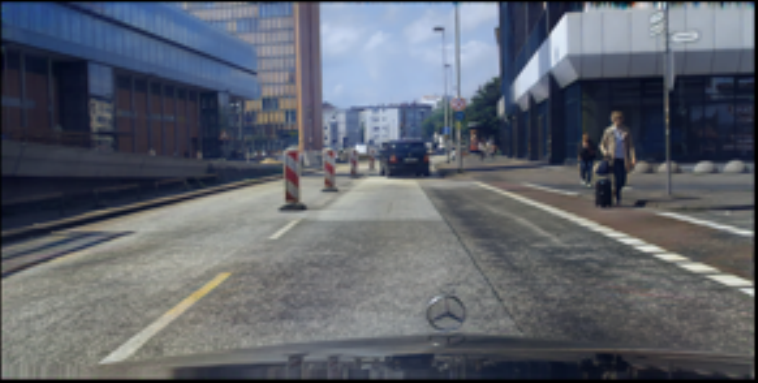}
    \includegraphics[width=0.2\linewidth]{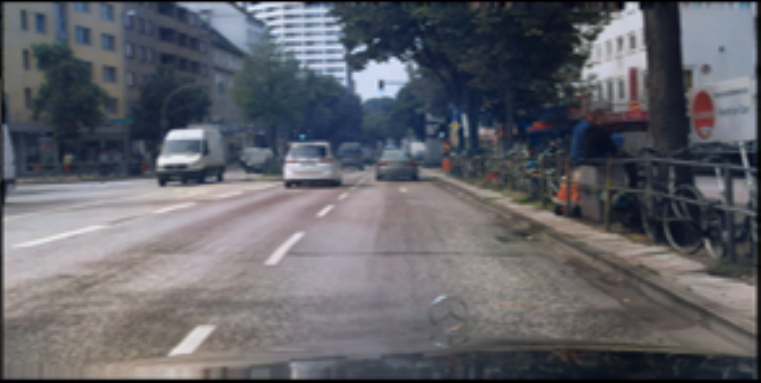}
    \includegraphics[width=0.2\linewidth]{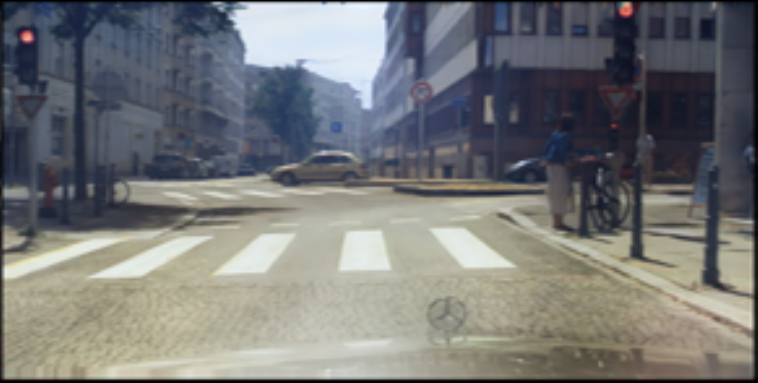}\\
    \small{(d) Transferred images using (c) CityScapes content and (a) GTA5 style.}\\
    \end{tabular}
	\caption{Domain adaptation results from our single DRANet in driving scenes. (a), (c) Original images. (b), (d) Transferred images.}
	\vspace{-2mm}
	\label{fig:segadaptation}
\end{figure*}

\section{Experiments}
\label{sec:experiments}
We evaluate DRANet for unsupervised domain adaptation on digit classification in~\secref{sec:digit} and driving scene segmentation in~\secref{sec:driving}.
We compare our bi-/tri-directional domain transfer results against multiple state-of-the-art un-/semi-supervised domain adaptation methods.
We also conduct an extensive ablation study to demonstrate the effectiveness of each proposed module in~\secref{sec:ablation}.
For the evaluations, we use the standard split of training and test sets the same as the existing unsupervised domain adaptations~\cite{bousmalis2017unsupervised, ye2020light}.
We train a task-classifier using stylized source training sets produced by DRANet and evaluate its performance on the target domain test sets.
We describe the training details in the supplementary materials.


\begin{table*}
\begin{center}
\begin{adjustbox}{max width = \textwidth}
\begin{tabular}{lcccccccccccccccccccccc}
\hline
& \begin{turn}{90}road\end{turn} & \begin{turn}{90}sidewalk\end{turn} & \begin{turn}{90}building\end{turn} & \begin{turn}{90}wall\end{turn} & \begin{turn}{90}fence\end{turn} & \begin{turn}{90}pole\end{turn} & \begin{turn}{90}traffic light\end{turn} & \begin{turn}{90}traffic sign\end{turn} & \begin{turn}{90}vegetation\end{turn} & \begin{turn}{90}terrain\end{turn} & \begin{turn}{90}sky\end{turn} & \begin{turn}{90}person\end{turn} & \begin{turn}{90}rider\end{turn} & \begin{turn}{90}car\end{turn} & \begin{turn}{90}truck\end{turn} & \begin{turn}{90}bus\end{turn} & \begin{turn}{90}train\end{turn} & \begin{turn}{90}motorbike\end{turn} & \begin{turn}{90}bicycle\end{turn} & \begin{turn}{90}\textbf{mIoU}\end{turn} & \begin{turn}{90}\textbf{fwIoU}\end{turn} & \begin{turn}{90}\textbf{Pixel Acc.}\end{turn}  \\
\hline
Source only & 42.7 & 26.3 & 51.7 & 5.5 & 6.8 & 13.8 & 23.6 & 6.9 & 75.5 & 11.5 & 36.8 & 49.3 & 0.9 & 46.7 & 3.4 & 5.0 & 0.0 & 5.0 & 1.4 & 21.7 & 47.4 & 62.5 \\
CyCADA~\cite{hoffman2018cycada} & 79.1 & 33.1 & 77.9 & 23.4 & 17.3 & 32.1 & \textbf{33.3} & \textbf{31.8} & 81.5 & 26.7 & 69.0 & 62.8 & 14.7 & 74.5 & 20.9 & 25.6 & 6.9 & 18.8 & 20.4 & 39.5 & 72.4 & 82.3 \\
LC~\cite{ye2020light} & 83.5 & 35.2 & 79.9 & 24.6 & 16.2 & \textbf{32.8} & 33.1 & \textbf{31.8} & 81.7 & 29.2 & 66.3 & \textbf{63.0} & 14.3 & \textbf{81.8} & 21.0 & 26.5 & 8.5 & 16.7 & \textbf{24.0} & 40.5 & 75.1 & 84.0 \\
Ours (without CADT)  & 83.5 & 33.7 & 80.7 & 22.7 & 19.5 & 25.2 & 28.6 & 25.8 & 84.1 & 32.8 & \textbf{84.4} & 53.3 & 13.6 & 75.7 & \textbf{21.7} & 30.6 & 15.8 & \textbf{20.3} & 19.5 & 40.6 & 75.6 & 84.9 \\
Ours (with CADT) & \textbf{85.0} & \textbf{35.8} & \textbf{82.0} & \textbf{26.4} & \textbf{21.6} & 27.0 & 29.2 & 28.1 & \textbf{84.2} & \textbf{34.0} & 81.9 & 53.6 & \textbf{15.9} & 73.6 & 21.1 & \textbf{31.0} & \textbf{16.7} & 17.2 & 22.8 & \textbf{41.4} & \textbf{76.4} & \textbf{85.7} \\
\hline
Target only & 97.3 & 79.8 & 88.6 & 32.5 & 48.2 & 56.3 & 63.6 & 73.3 & 89.0 & 58.9 & 93.0 & 78.2 & 55.2 & 92.2 & 45.0 & 67.3 & 39.6 & 49.9 & 73.6 & 67.4 & 89.6 & 94.3 \\
\hline
\end{tabular}
\end{adjustbox}
\end{center}
\caption{Result comparison of DRANet to state-of-the-art methods on domain adaptation for semantic segmentation. We also report the performance of DRANet with and without Content-Adaptive Domain Transfer (CADT).}
\label{tab:segmentation}
\end{table*}

\begin{figure*}[t] 
	\centering
	\begin{tabular}{c@{\hspace{1mm}}c@{\hspace{1mm}}c@{\hspace{1mm}}c@{\hspace{1mm}}}
    \includegraphics[width=0.2\linewidth]{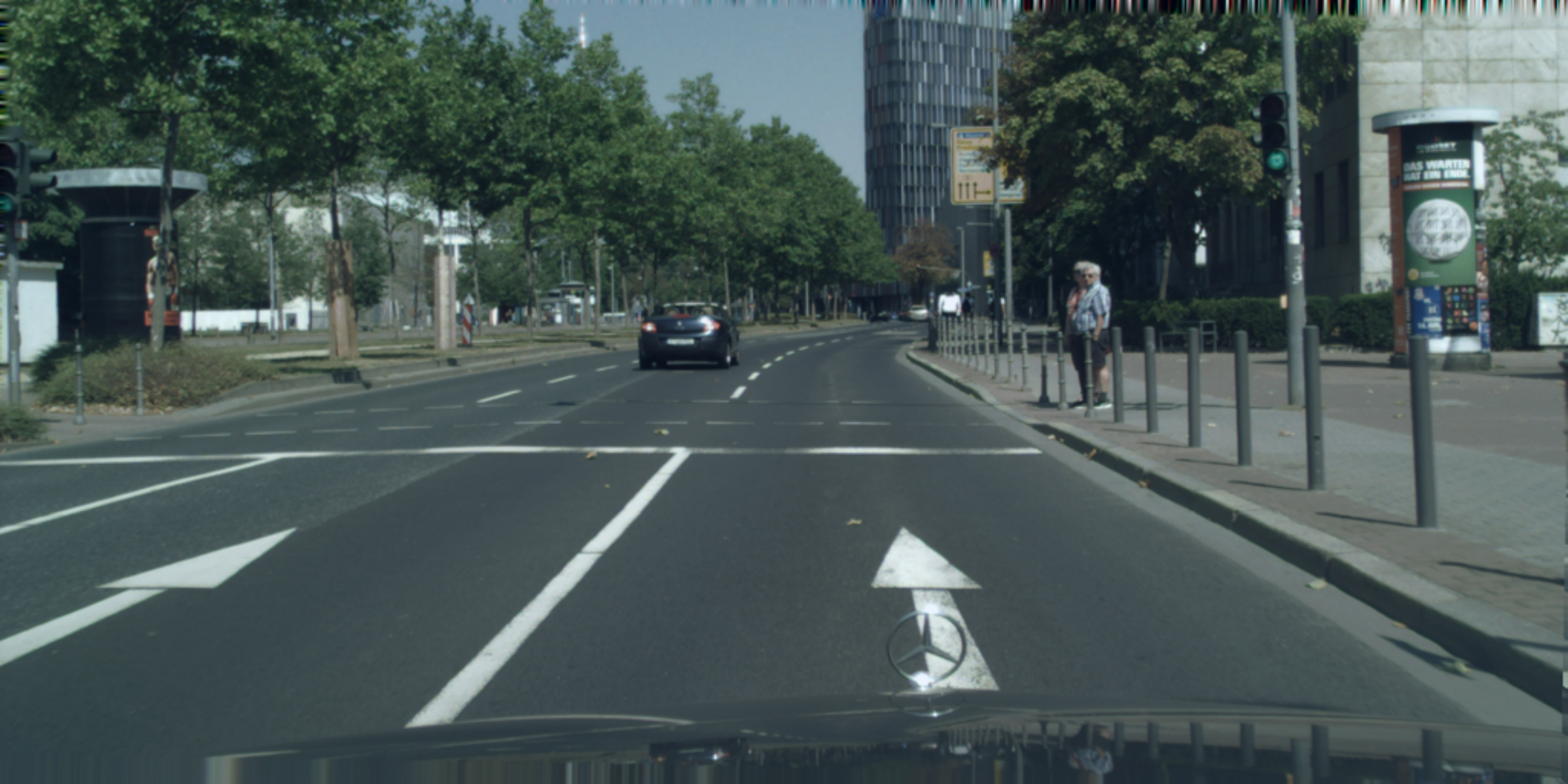}
    \includegraphics[width=0.2\linewidth]{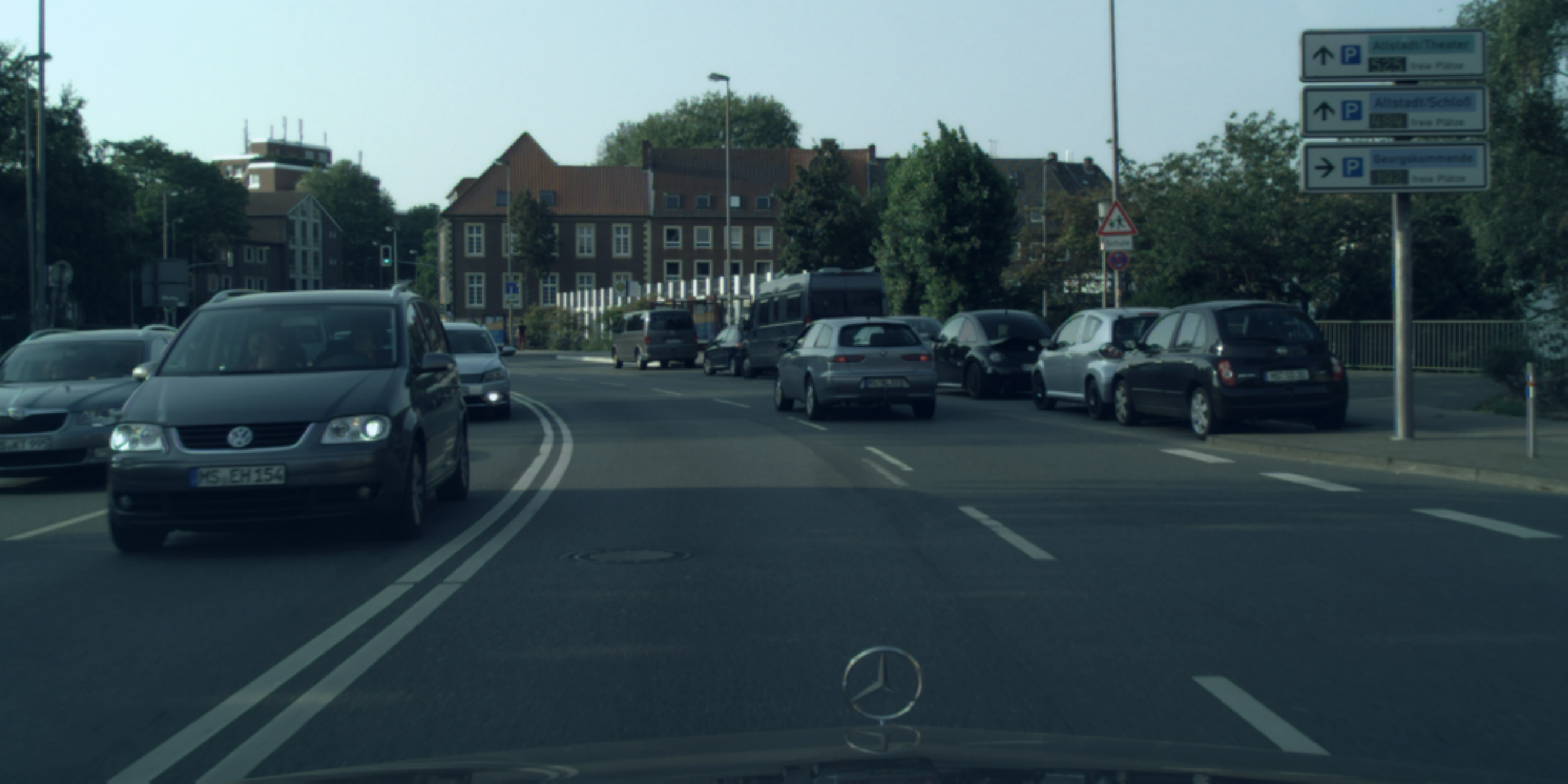}
    \includegraphics[width=0.2\linewidth]{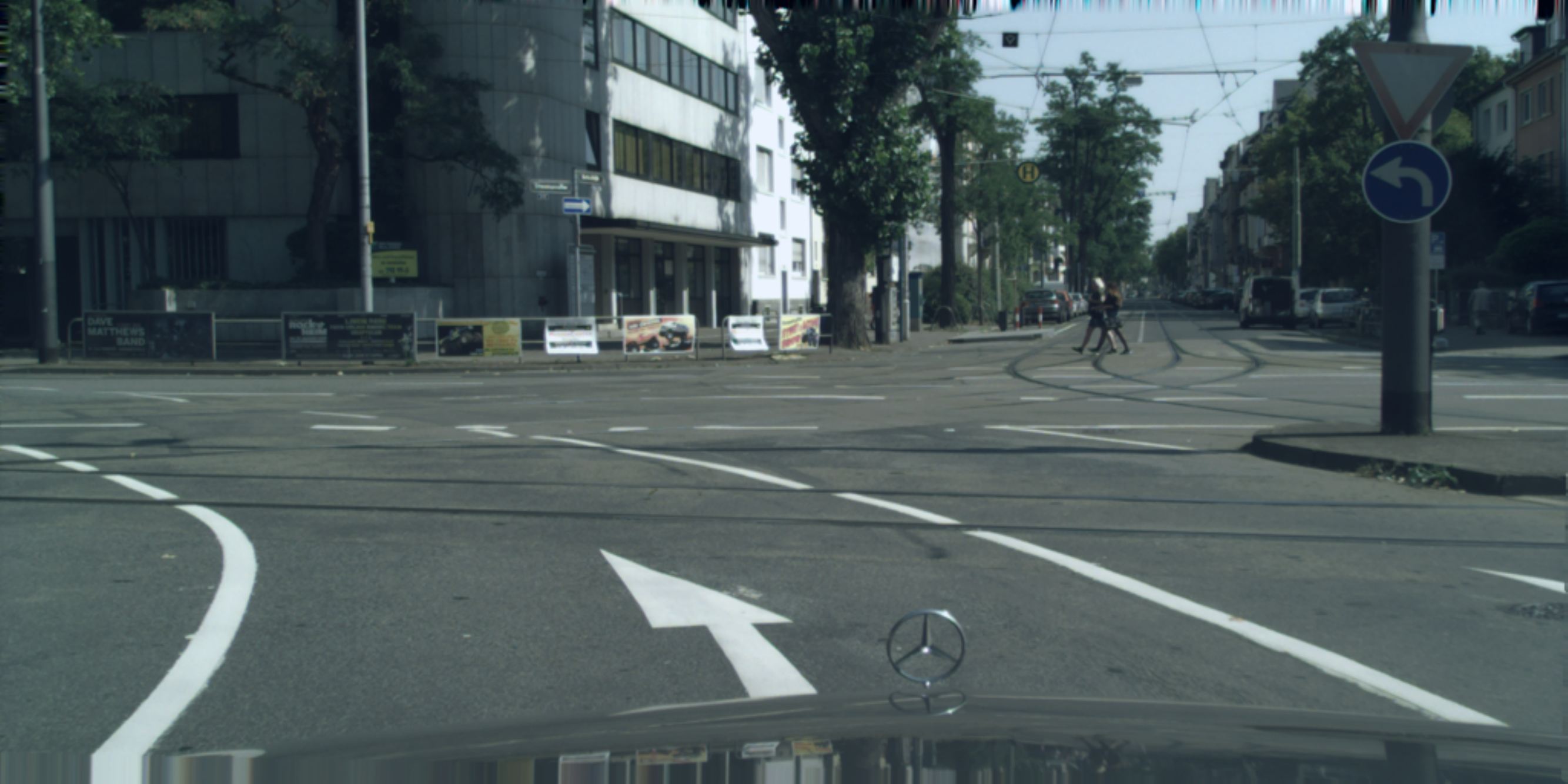}
    \includegraphics[width=0.2\linewidth]{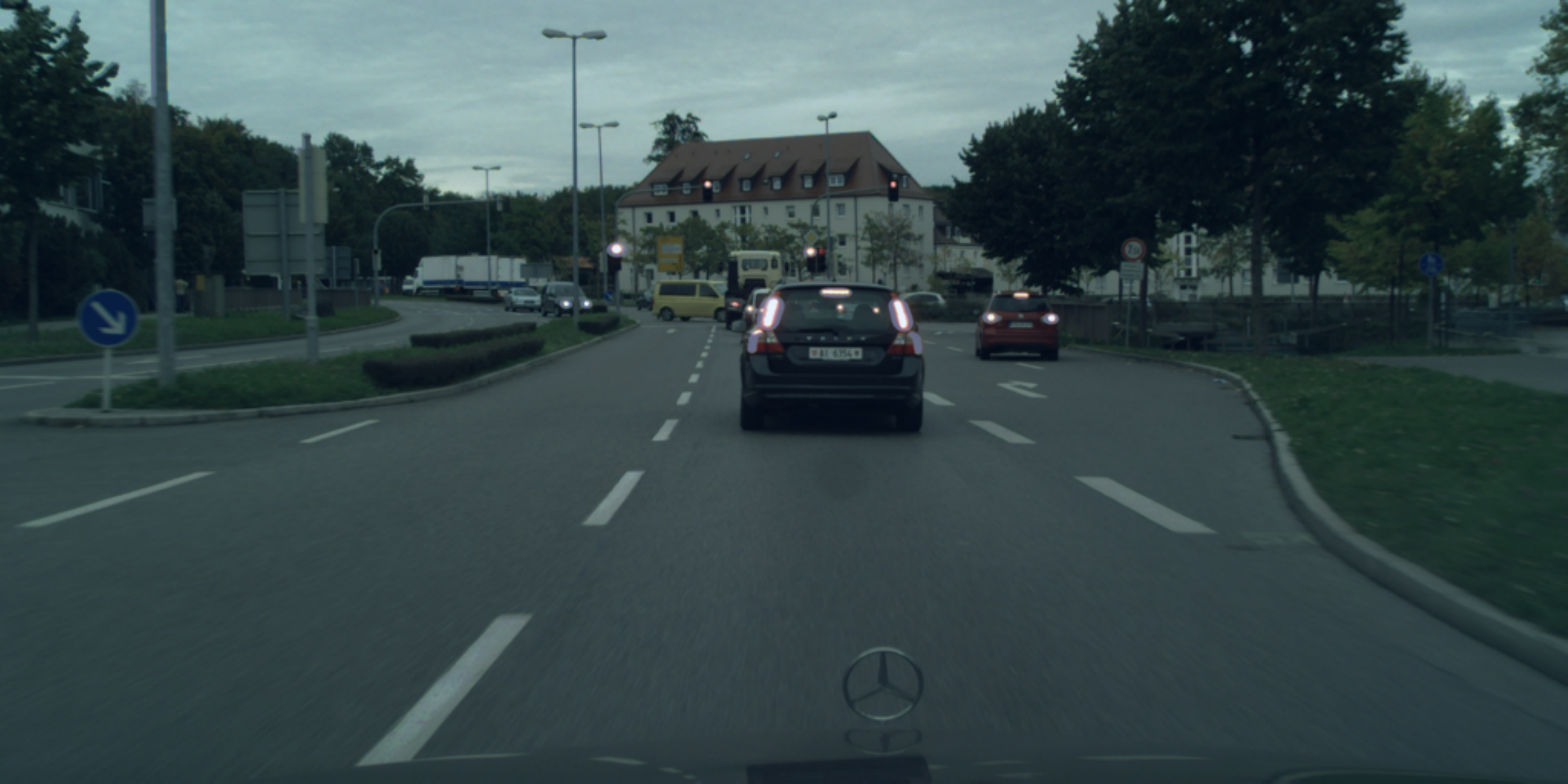}\\
    \small{(a) Test images (CityScapes)}\\
    \includegraphics[width=0.2\linewidth]{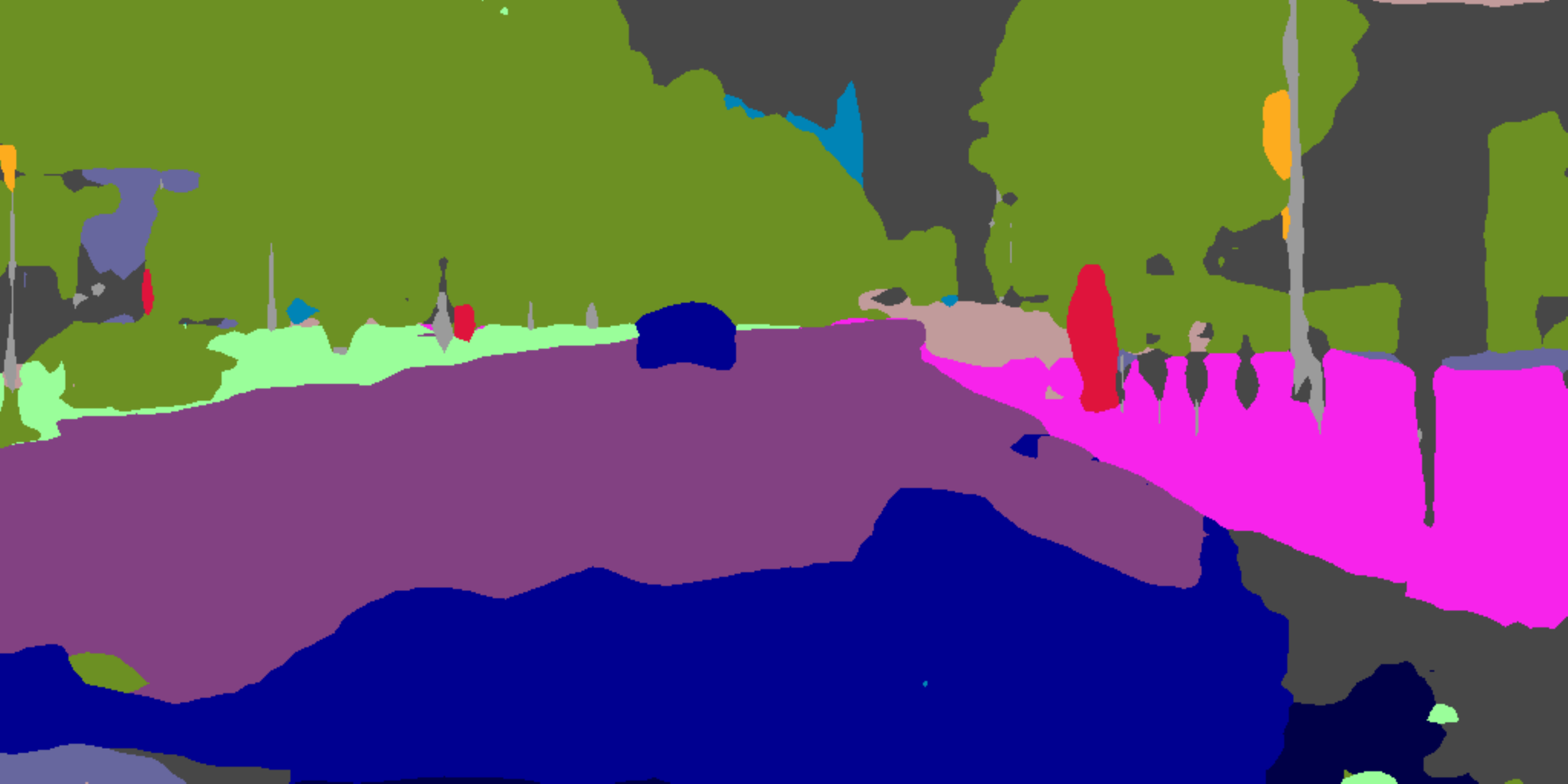}
    \includegraphics[width=0.2\linewidth]{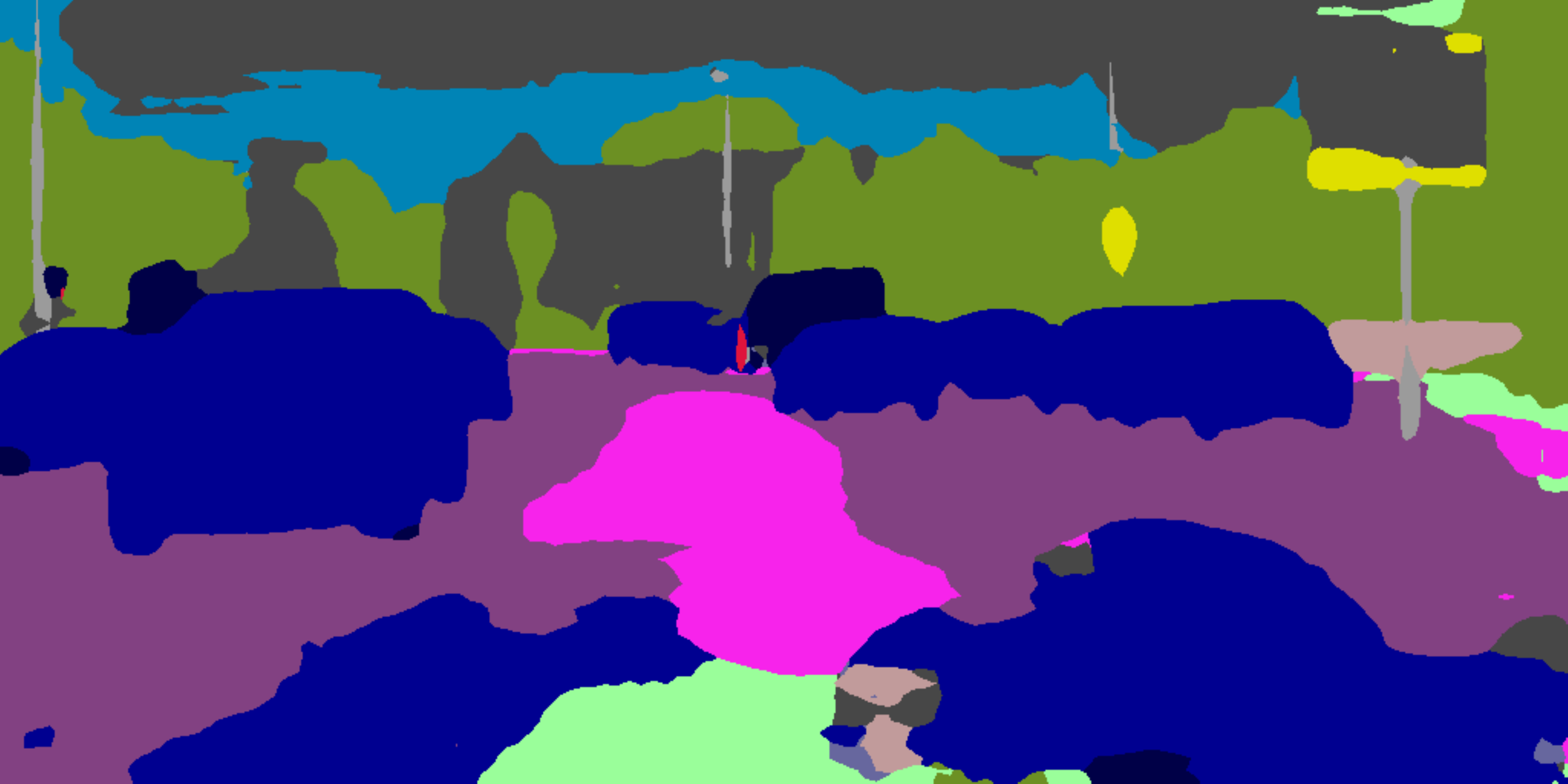}
    \includegraphics[width=0.2\linewidth]{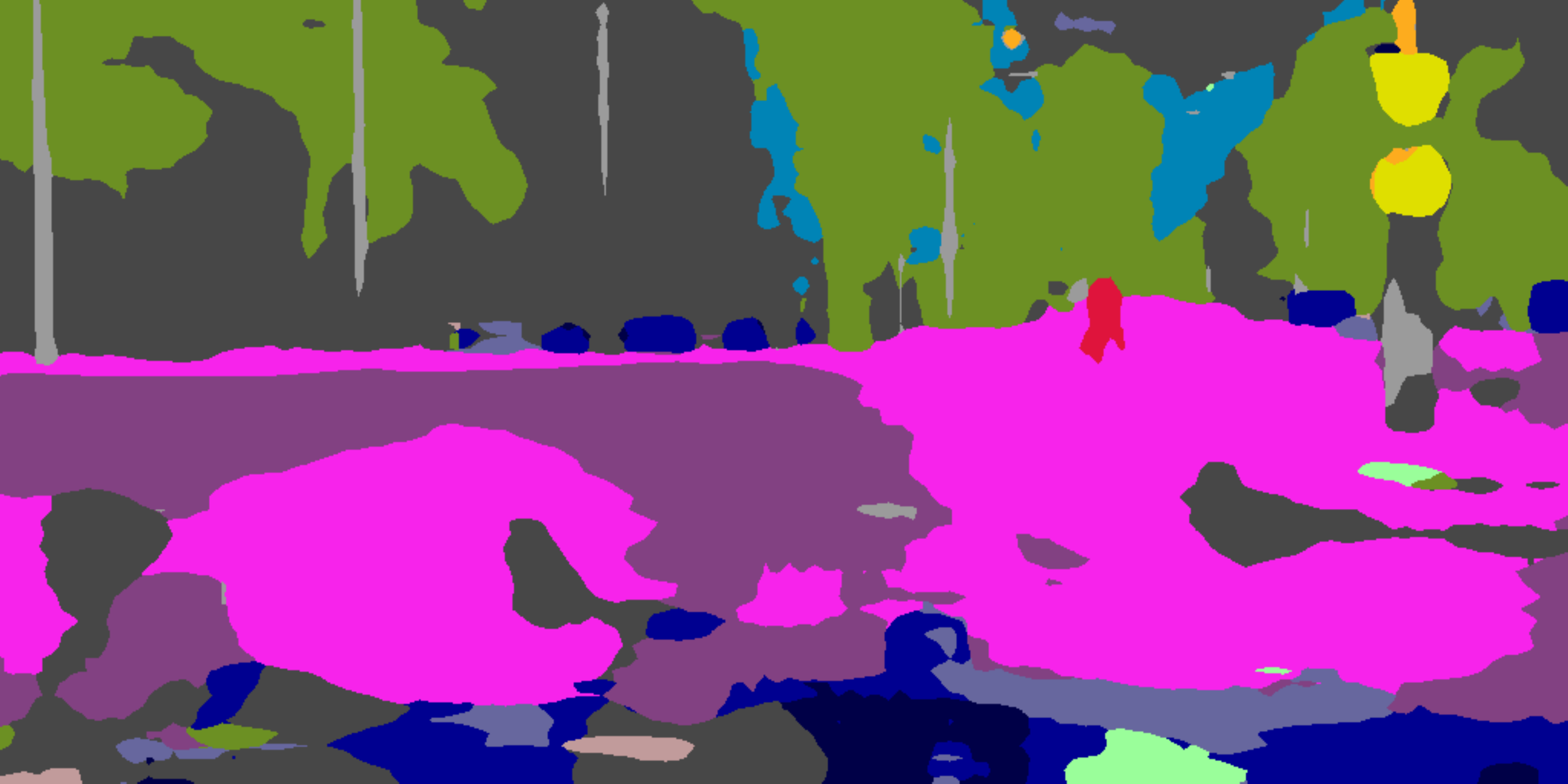}
    \includegraphics[width=0.2\linewidth]{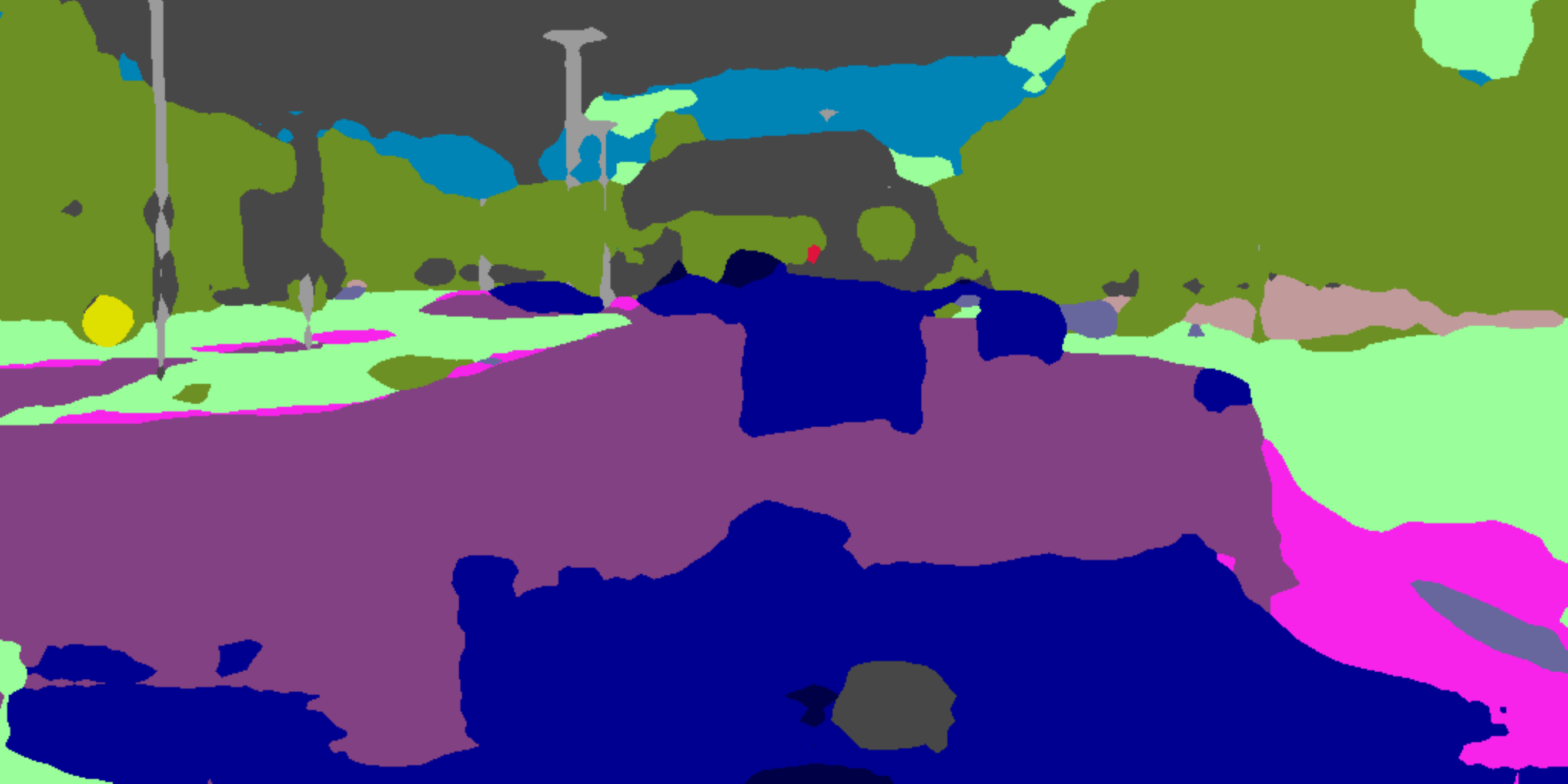}\\
    \small{(b) Source prediction.}\\
    \includegraphics[width=0.2\linewidth]{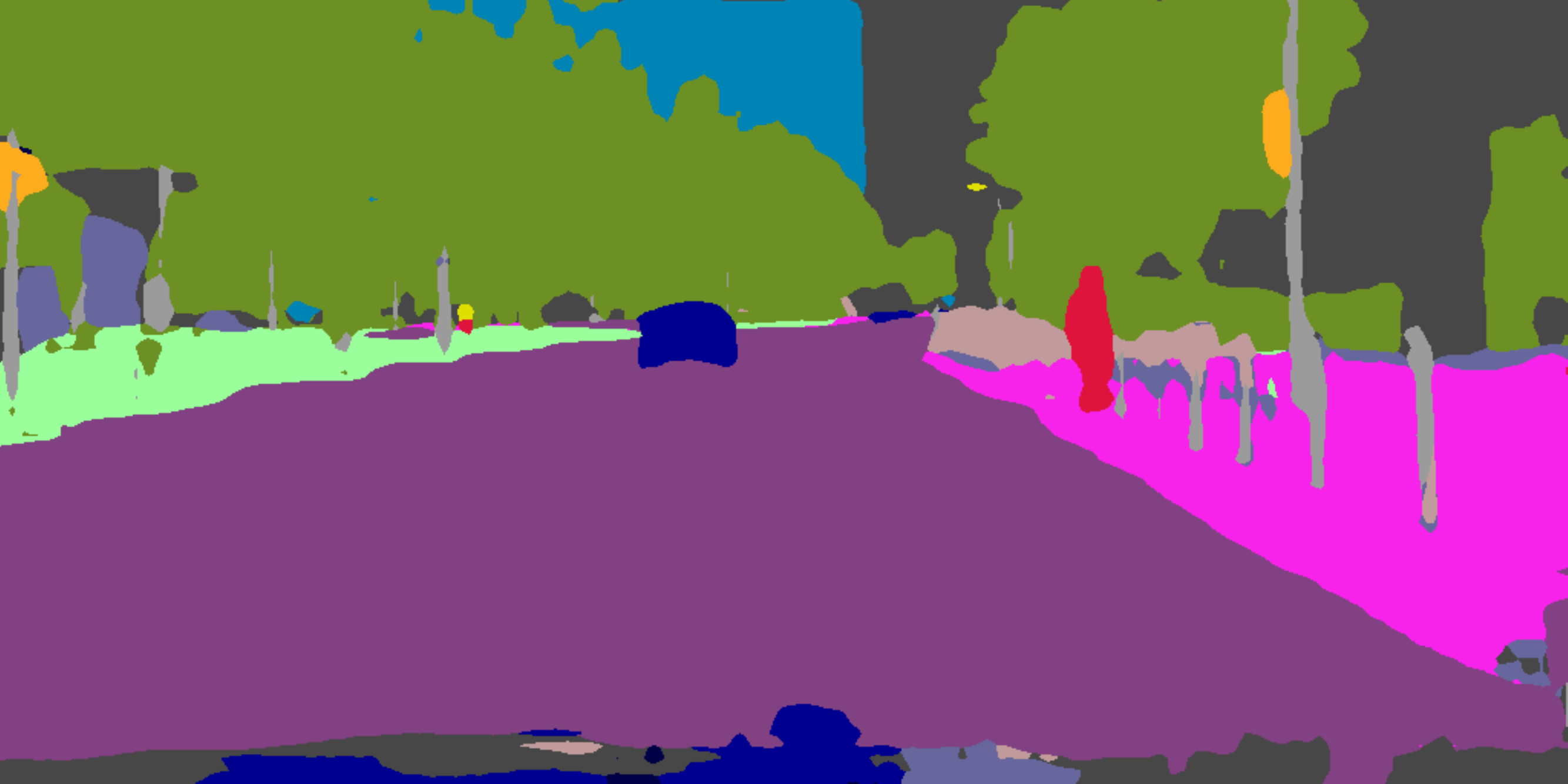}
    \includegraphics[width=0.2\linewidth]{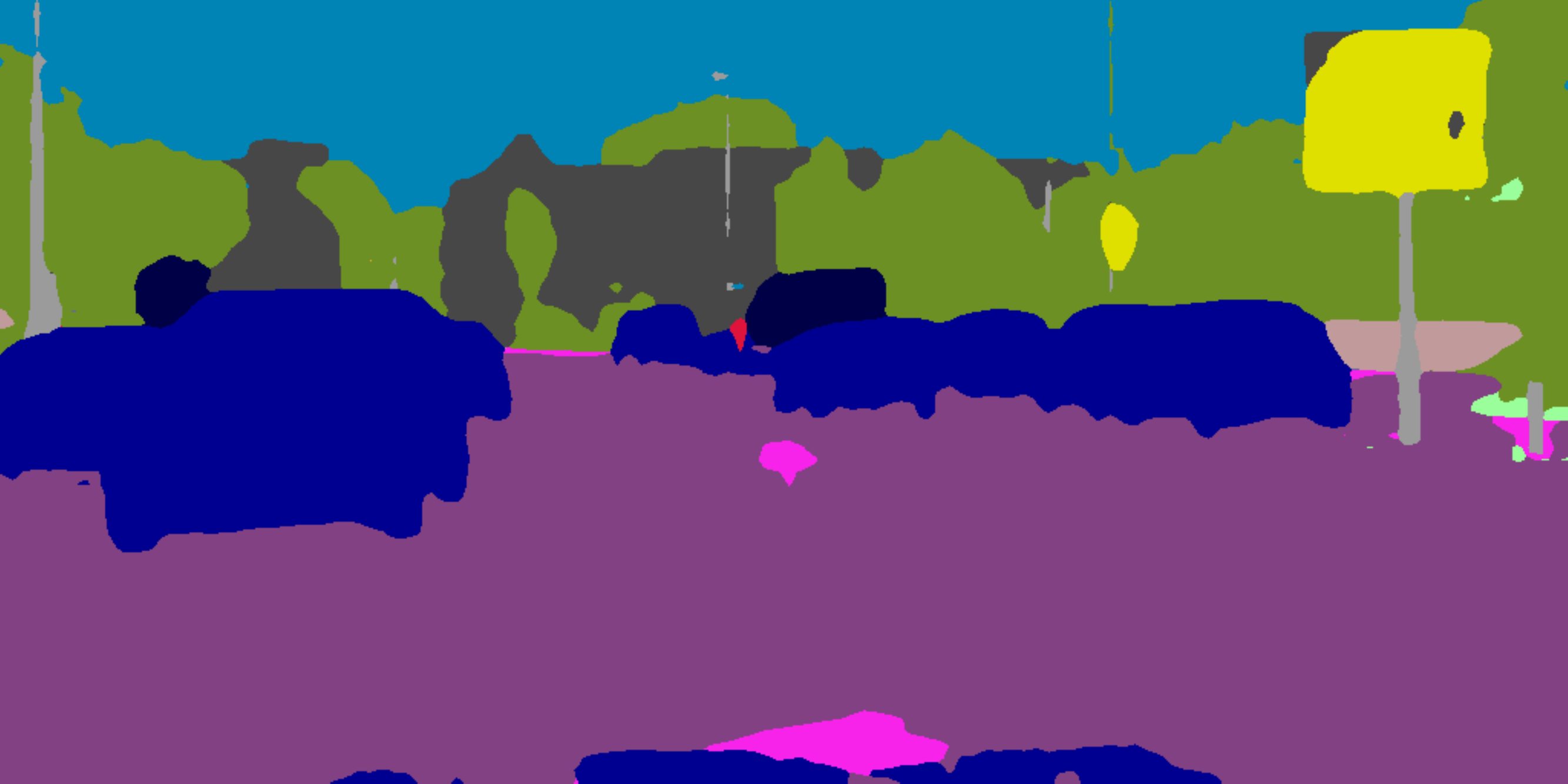}
    \includegraphics[width=0.2\linewidth]{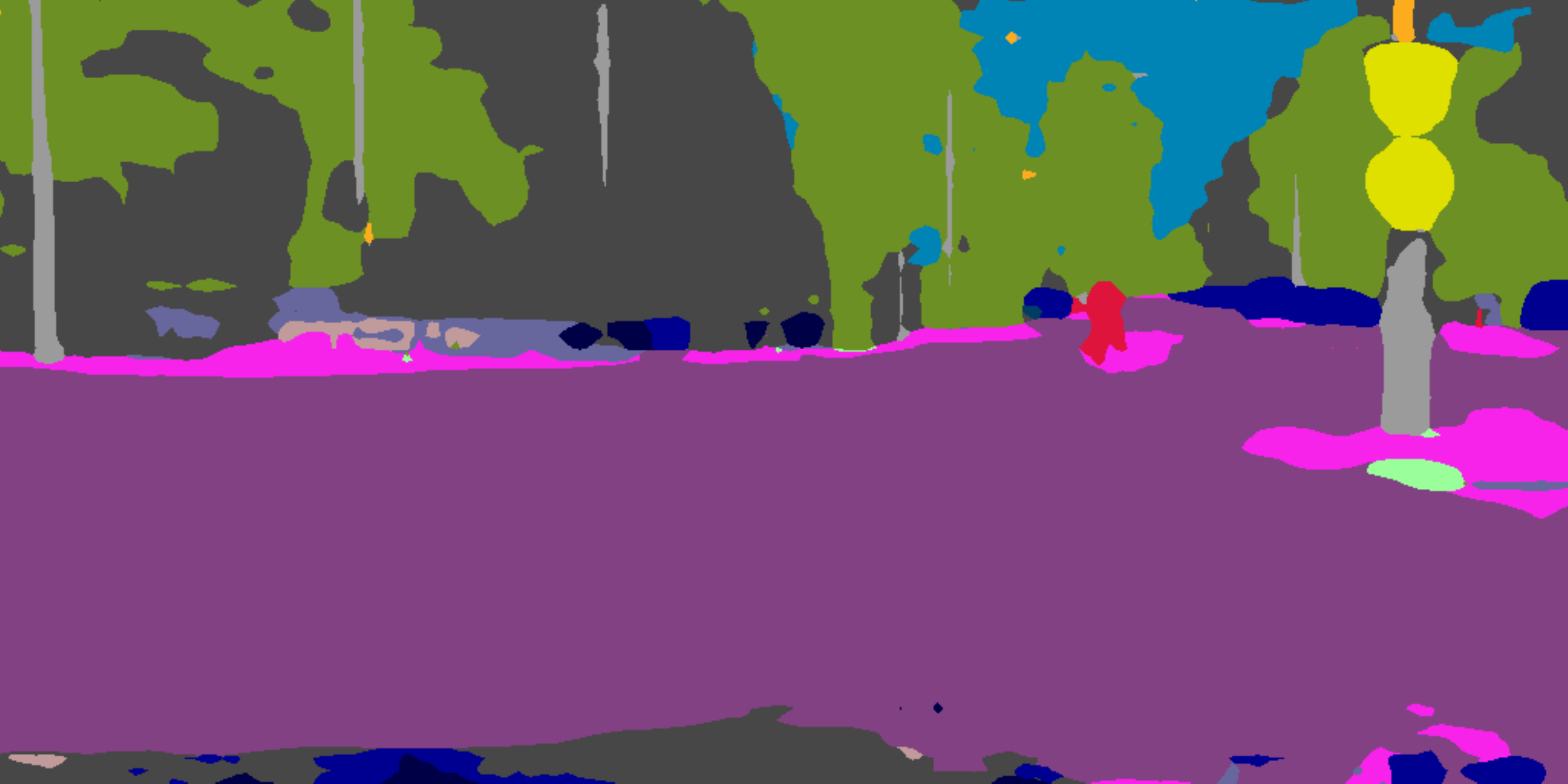}
    \includegraphics[width=0.2\linewidth]{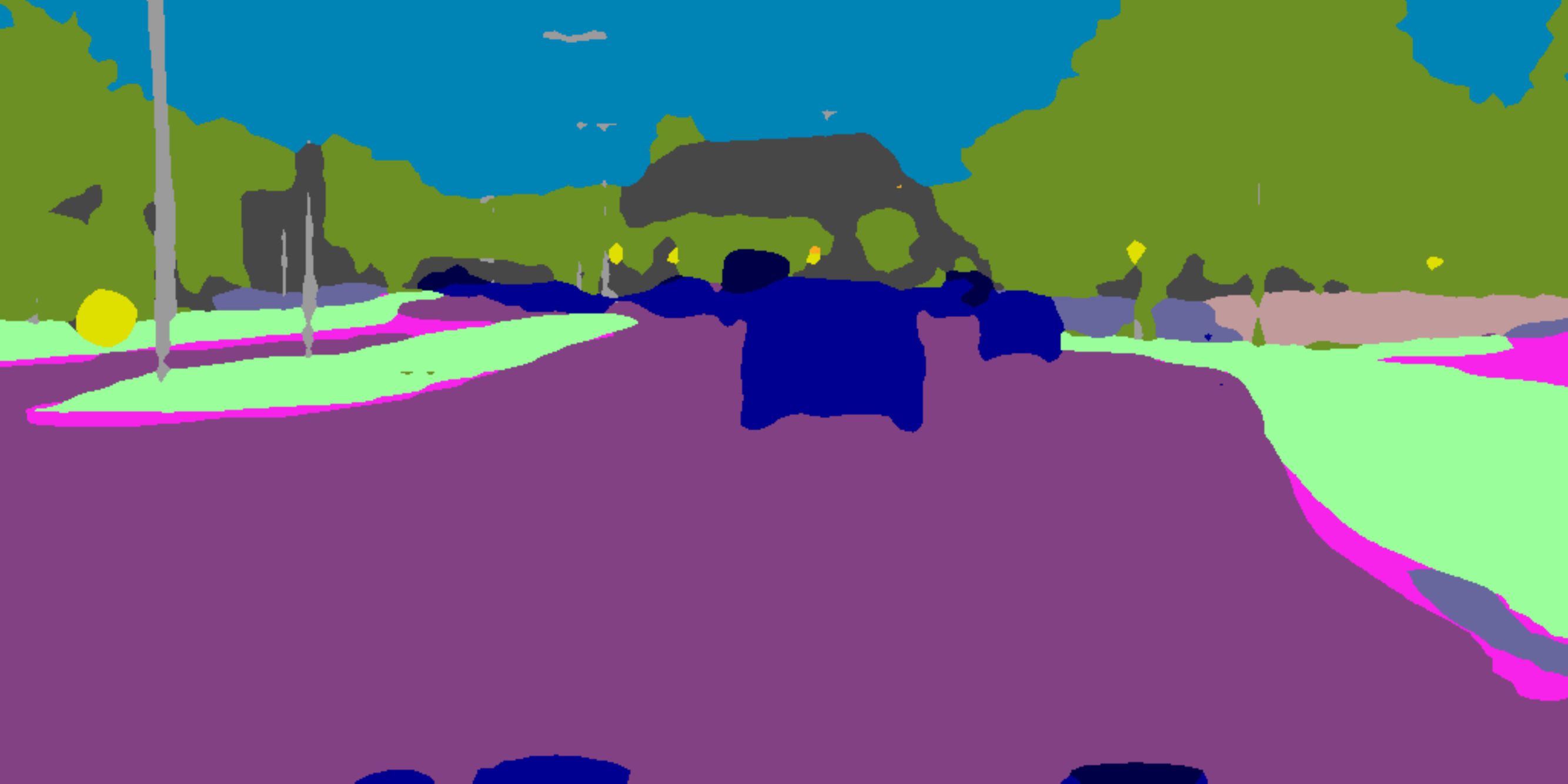}\\
    \small{(c) Our prediction}\\
    \includegraphics[width=0.2\linewidth]{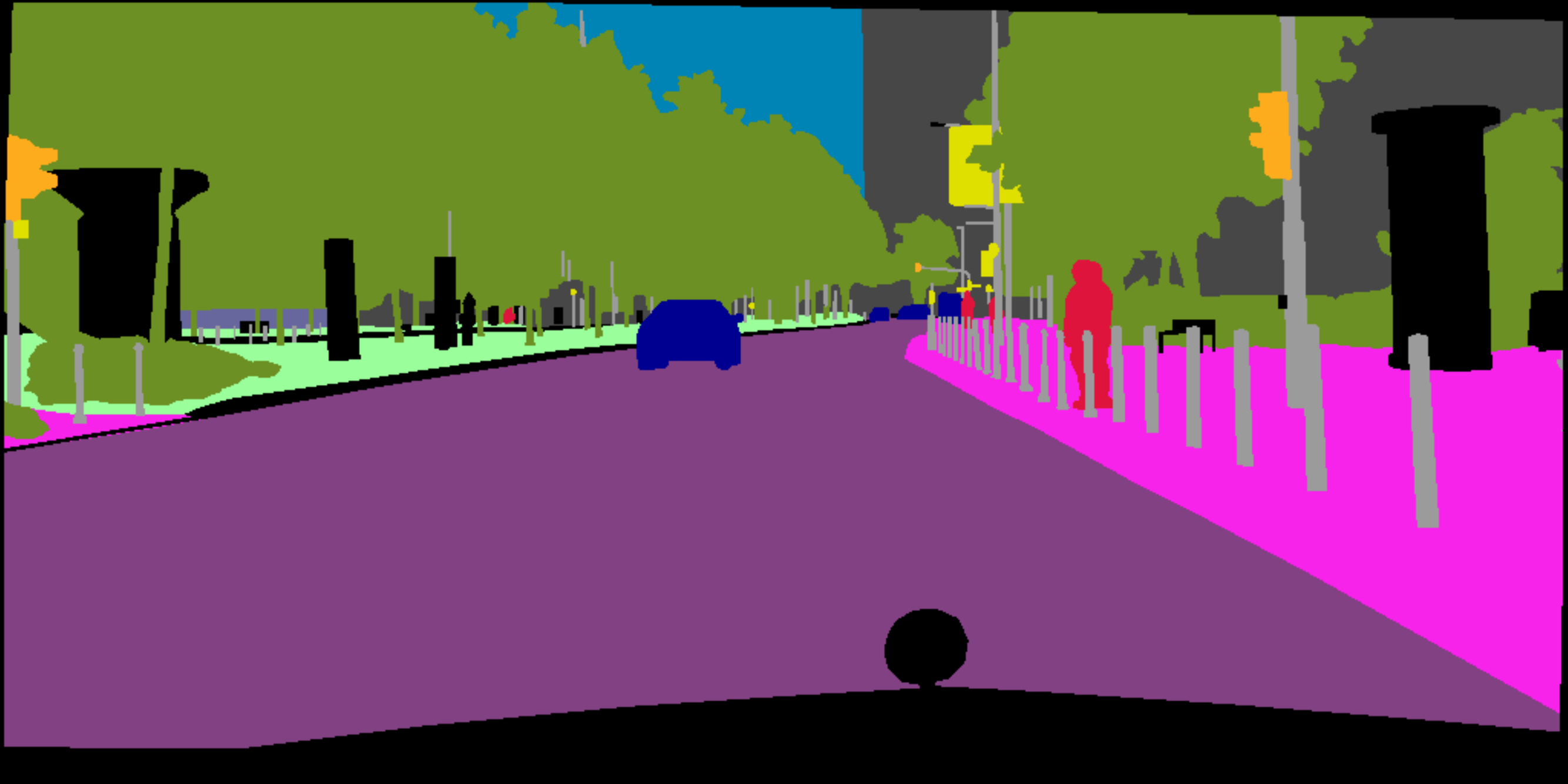}
    \includegraphics[width=0.2\linewidth]{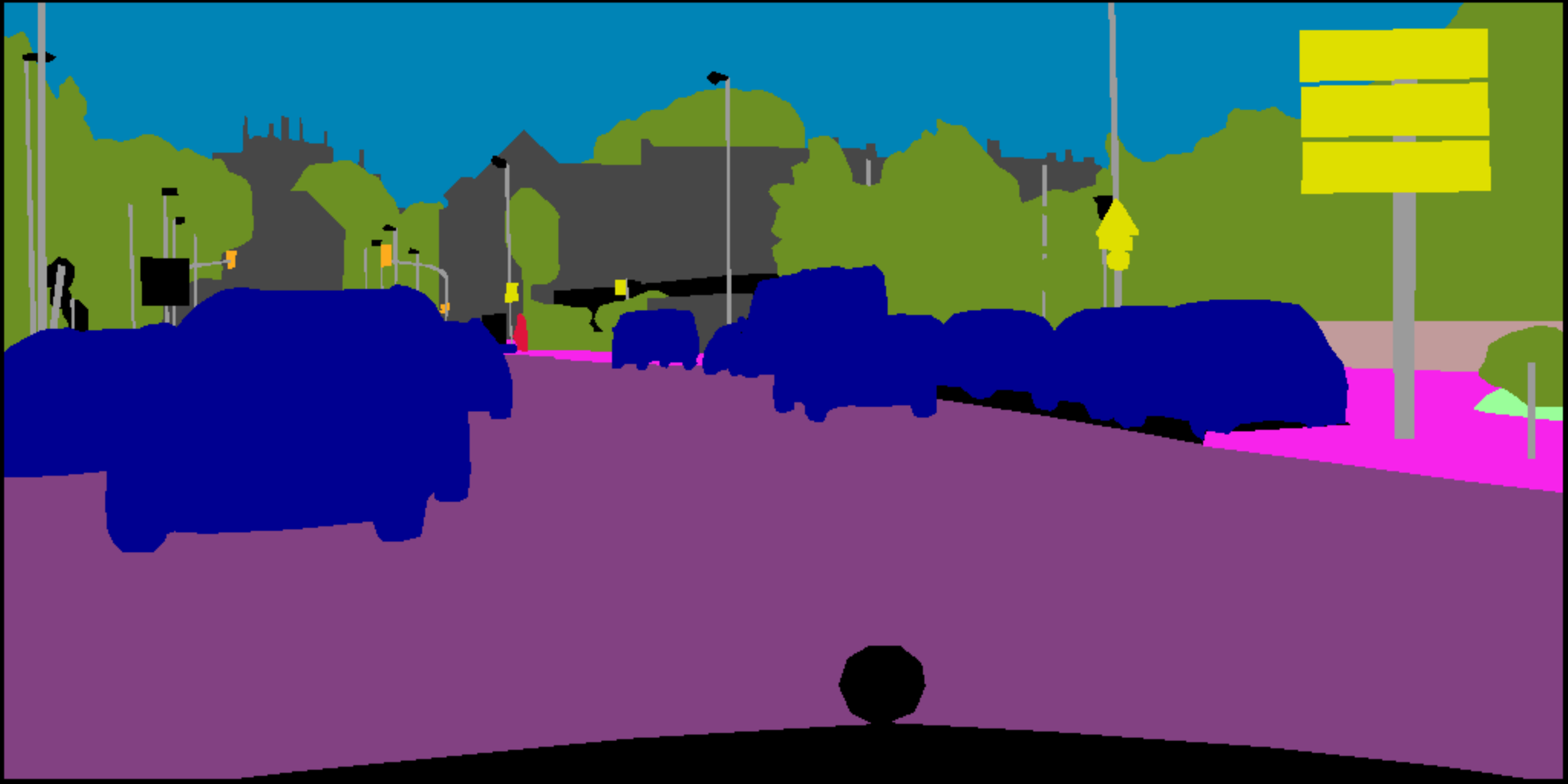}
    \includegraphics[width=0.2\linewidth]{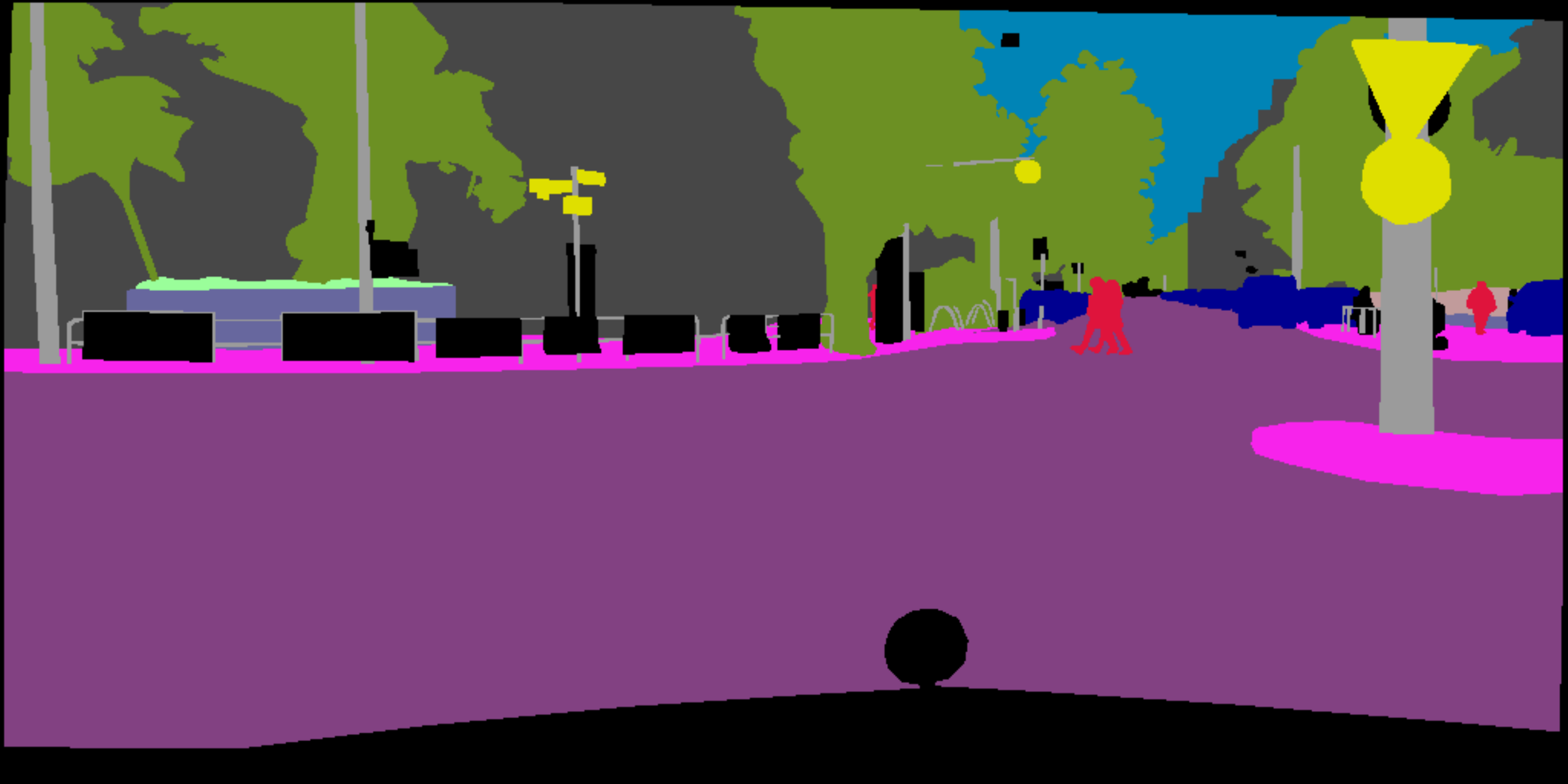}
    \includegraphics[width=0.2\linewidth]{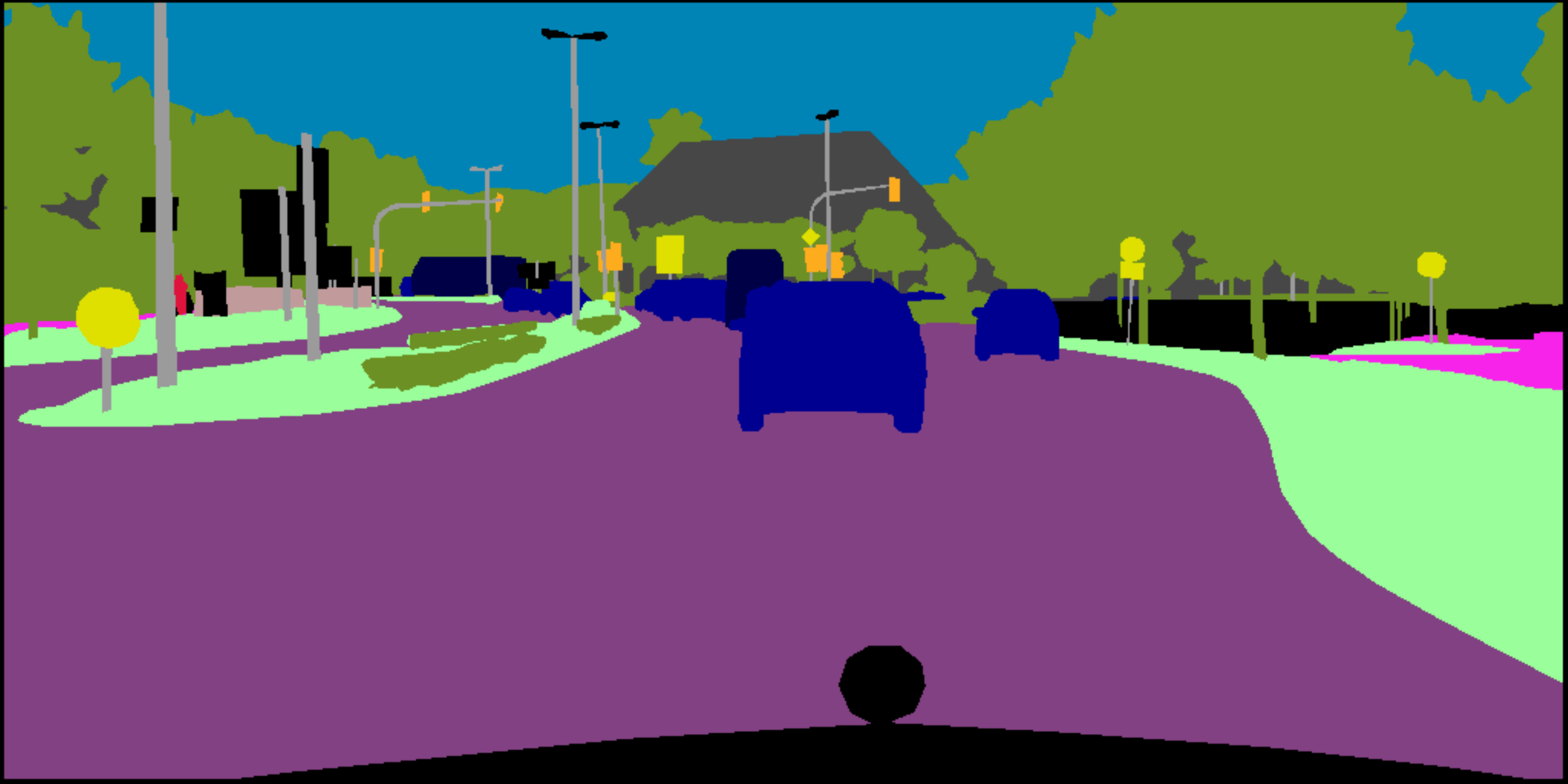}\\
    \small{(d) Ground Truth}\\
	\end{tabular}
	\caption{Semantic segmentation results for GTA5 to CityScapes. Note that we do not use any GT segmentation labels for training DRANet. }
	\label{fig:segresult}
	\vspace{-3mm}
\end{figure*}

\subsection{Adaptation for Digit Classification}
\label{sec:digit}
Unlike existing domain adaptation methods, where a single model is responsible for domain transfer in one direction, our single model is able to deal with multi-directional domain adaption.
We demonstrate the versatility of DRANet by transferring images across the multiple domains using three digit datasets: MNIST \cite{lecun1998gradient}, MNIST-M \cite{ganin2016domain}, and USPS \cite{hull1994database}.
We train our model for bi-directional domain adaptation (MNIST to MNIST-M or USPS, and its opposite direction) as shown in~\figref{fig:overview}. 
We also train the adaptation model tri-directionally (MNIST to MNIST-M and USPS, and their opposite directions) and show the results in~\figref{fig:3domain}.
Note that we have not explicitly transferred the domain between MNIST-M and USPS during training, but the results show that DRANet is also applicable for the adaptation between them.

As shown in~\tabref{tab:digit}, our model, either trained for two or three domains, outperforms all the competitive methods~\cite{ye2020light, hoffman2018cycada, bousmalis2017unsupervised, liu2016coupled, tzeng2017adversarial, bousmalis2016domain, ganin2016domain}.
The results also show that our model even achieves higher performance than the model trained only on target except for the experiment of USPS to MNIST.
This is because DRANet augments as many images as the number of target images using one source image as shown in~\figref{fig:transferred}.
DRANet-based data augmentation makes the classifier even more robust than the target-only model.
Moreover, we show the content similarity matrix in~\figref{fig:confusion} that reveals how well our model disentangles the representation into content and style components.
We use 10 images with similar content from MNIST and MNIST-M each, and observe the confusion matrix has the highest diagonal values.
We also observe that both higher values around $50\%$ for both samples of digit one.
The results show that our model disentangles the representation of content and style while maintaining each domain's characteristics.

\subsection{Adaptation for Semantic Segmentation}
\label{sec:driving}
To show the applicability of DRANet on the complex real-world scenario, we use GTA5 \cite{richter2016playing} and Cityscapes \cite{cordts2016cityscapes}, which contain driving scene images with dense annotations.
We train our model using 24966 images in GTA5 and 2975 images in Cityscapes train set, and we train DRN-26 \cite{yu2017dilated} with 19 common classes for synthetic to real adaptation.
The results in~\figref{fig:segadaptation} show that our model generates stylized images following the artistic appearance of target images while keeping the scene structure of source images.
We also evaluate the domain adaptation performance on semantic segmentation. 
The quantitative results in~\tabref{tab:segmentation} show that our model achieves state-of-the-art performance in all three main metrics for semantic segmentation: mIoU, fwIoU, and pixel accuracy.
Among the 19 segmentation labels, our method outperforms the competitive methods in 14 categories.
Especially, the accuracy of sky labels is improved by a large margin.
We believe that our model designed for maintaining the scene structure allows to stably generate domain transferred images as shown in~\figref{fig:segadaptation}, and leads the performance improvement as shown in~\figref{fig:segresult}.
 

\begin{figure*}[t] 
	\centering
	\begin{tabular}{c@{\hspace{1mm}}c@{\hspace{1mm}}c@{\hspace{1mm}}}
    \includegraphics[width=0.29\linewidth]{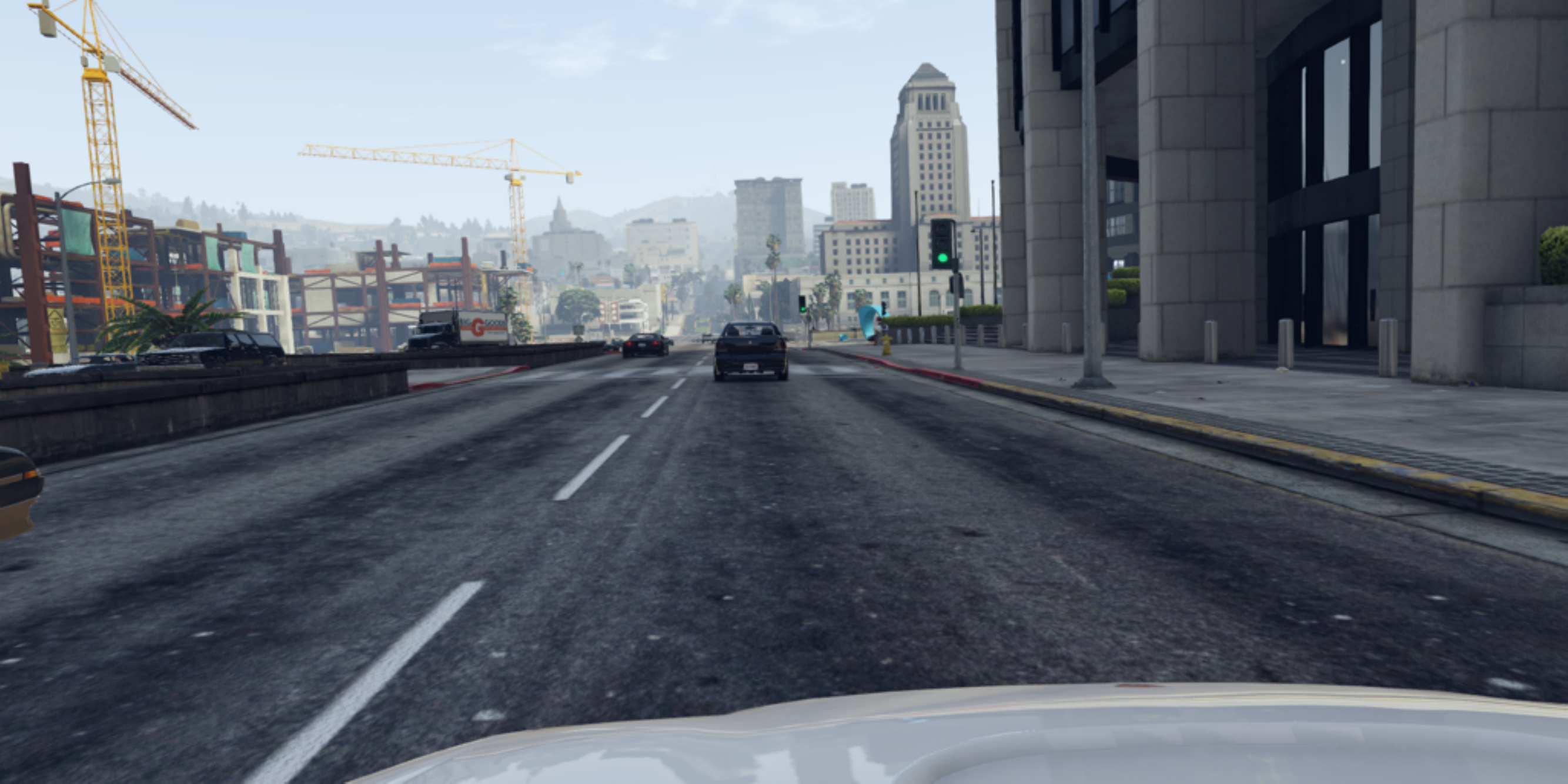} &
    \includegraphics[width=0.29\linewidth]{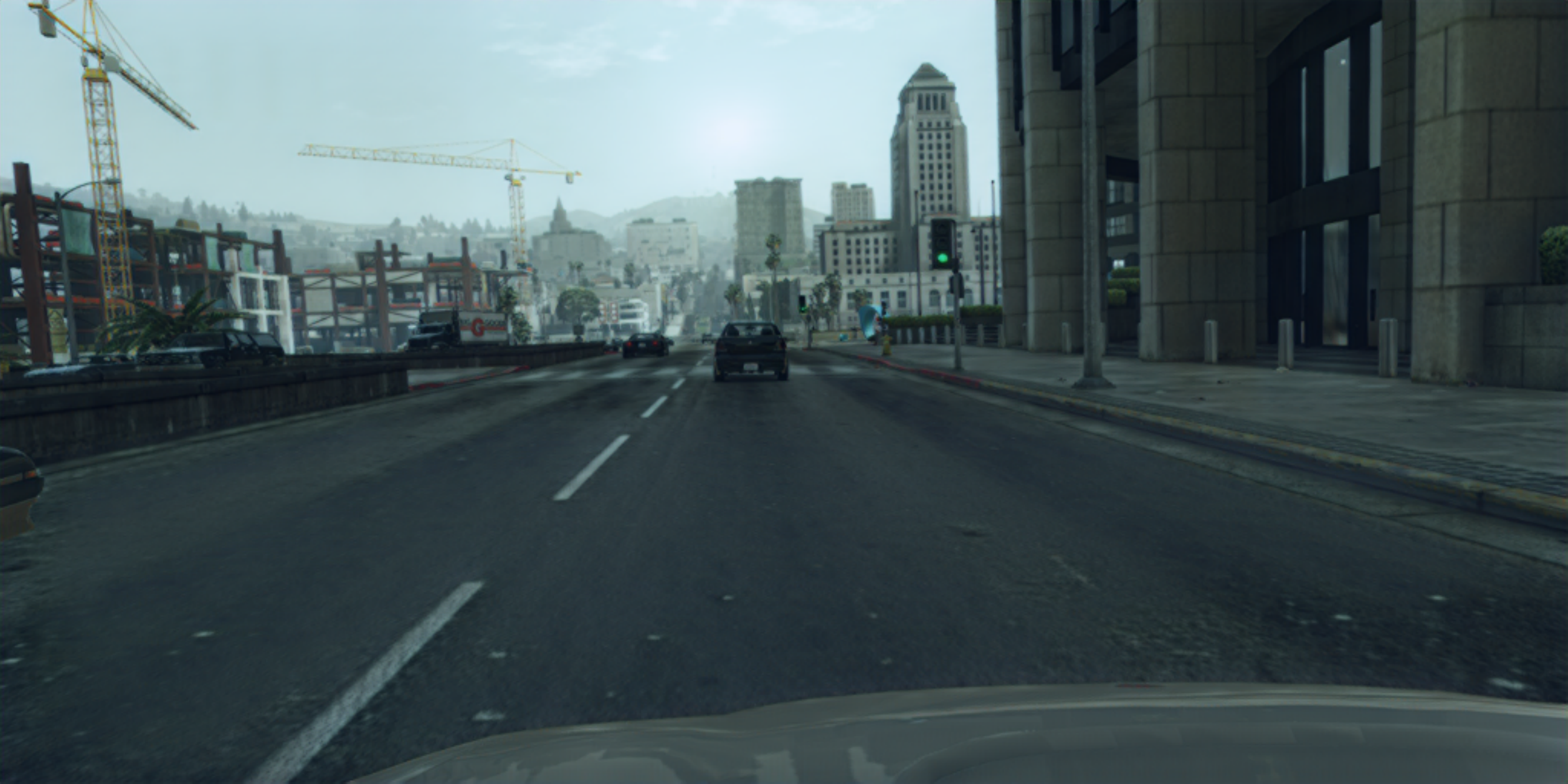} &
    \includegraphics[width=0.29\linewidth]{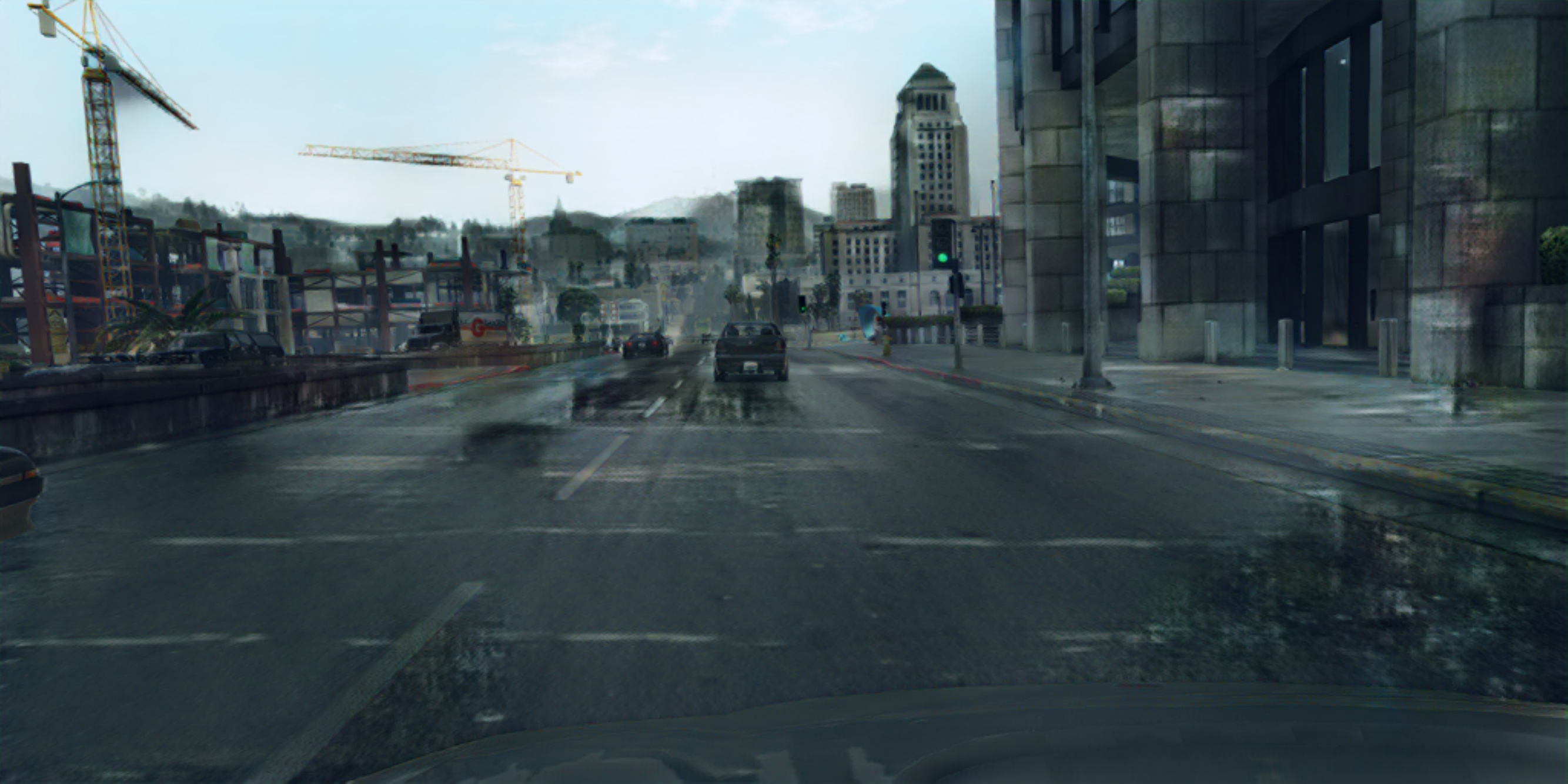}  \\
    \small{(a) Source image (GTA5)} &  \small{(b) Ours: CADT using all image in (d)} & \small{(c) DT using the rightmost image in (d)}\\
    \end{tabular}
	\begin{tabular}{c@{\hspace{1mm}}}
    \includegraphics[width=0.88\linewidth]{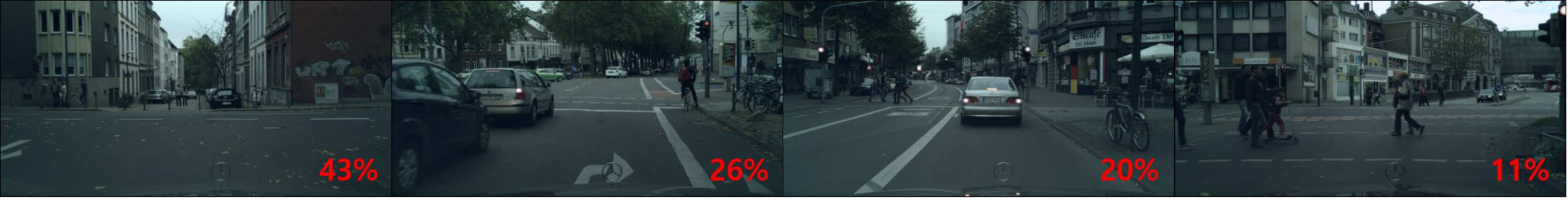}\\
    \small{(d) Target images (CityScapes) in a minibatch and their content similarity with the source image in (a)}\\
     \end{tabular}
	\caption{Comparison on image synthesis using Content-Adaptive Domain Transfer (CADT) and normal Domain Transfer (DT).}
	\label{fig:contentadaptive}
	\vspace{-3mm}
\end{figure*}

\subsection{Ablation Study}
\label{sec:ablation}
\textbf{Representation Disentangler} \quad
We design our separator incorporating two key ideas: one is the non-linearity of feature mapping and the other is domain normalization factor.
To show the effectiveness of these key contributions, we set four experiment settings with/without non-linearity and normalization factors in our framework.
We evaluate DRANet in each set for two bidirectional domain transfer tasks (one between MNIST and USPS, the other between MNIST and MNIST-M).
We compare the classification results of each case in unsupervised domain adaptation.
As shown in~\tabref{tab:ablation}, our model involving both non-linearity and normalization factors shows the best performance among four different settings.
In the adaptation task between MNIST and MNIST-M, all model, even without non-linearity and normalization factor, produces the reasonable performance because both datasets contain the same content representation.
Note that MNIST-M is one variation on MNIST proposed for unsupervised domain adaptation, which replaces the background of images while maintaining each MNIST digit~\cite{ganin2016domain}.
However, there is a large gap in each case for adaptation between MNIST and USPS, which have obviously different content representation.
The model without both components results in poor classification performance of one side. This means the model can only adapt either directional domain adaptation (MNIST to USPS or USPS to MNIST), like what the existing methods do.
The model with either non-linearity or normalization improves the performance while our model with both factors achieves the best among the other settings.
We empirically demonstrate that non-linear mapping affords better representation disentanglement and the drastic performance improvement.
As the advantages of non-linear mapping function of features proven in~\cite{scholkopf1997kernel}, we believe that the non-linearity is considerably responsible for clear separation of representations.
We also show the normalization factor further boosts the adaptation performance. 
We can conclude that both factors play an important role in representation disentanglement as well as in an unsupervised domain adaptation.

\begin{table}[t]
\renewcommand{\tabcolsep}{1mm}
\centering
\begin{tabular}{@{}cc|cccccc@{}}
\hline
\small{Non-}  & \small{Normal-}  & \small{MNIST} & \small{USPS to} & \small{MNIST to} & \small{MNIST-M} \\ 
\small{linearity} & \small{ization} & \small{to USPS} & \small{MNIST}  & \small{MNIST-M} & \small{to MNIST} \\
\hline
 &  & 11.2 & 87.1 & 97.0 & 99.0 \\
& \checkmark & 90.7 & 90.2 & 97.7 & 99.1  \\
\checkmark & & 96.6 & 90.9 & 97.3 & 98.9   \\ 
\checkmark & \checkmark & \textbf{98.2} & \textbf{97.8} & \textbf{98.7} & \textbf{99.3} \\
\hline
\end{tabular}
\caption{Ablation study on the separator design to verify the effectiveness of the non-linearity in representation disentanglement (Non-linearity) and distribution scale parameters (Normalization).}
\vspace{-5mm}
\label{tab:ablation}
\end{table}

\textbf{Content-Adaptive Domain Transfer} \quad
This subsection shows two advantages of our CADT for domain adaptation.
One is that it prevents the model to be trained with bad training samples and the other is that it encourages the model to generate better-stylized images.
During the early phase of training, the separator is not able to clearly disentangle the content and style components, which means each separation does not solely involve its identical information.
Consequently, the model generates content-mixed images at the early training stage, and it might disturb the training by fooling discriminator, especially in the case the two images have quite different content.
These strengths can be observed in~\figref{fig:contentadaptive} that shows the comparison of the results from the model with/without CADT trained with less than 1000 iterations.
\figref{fig:contentadaptive}-(a) is a source image (GTA5), and \figref{fig:contentadaptive}-(d) contains multiple target images (Cityscapes) in one minibatch. The bottom-right digit indicates the content similarity with source image.
We show the domain transferred images with/without CADT in \figref{fig:contentadaptive}-(b),(c), respectively.
The result in \figref{fig:contentadaptive}-(c) is generated by adapting the domain of rightmost target image in \figref{fig:contentadaptive}-(d), which has the lowest similarity.
The results show the normal domain transfer causes the significant artifact in the early stage of training while the proposed CADT reasonably synthesizes the image.
It means that CADT helps to disentangle the representation even at just a few iterations.
We also show that the general performance improvement by CADT in~\tabref{tab:segmentation} by comparing the domain adaptation results with/without CADT.
The table demonstrates the effectiveness of our content-adaptive domain transfer.

\section{Conclusion}
\label{sec:conclusion}
In this paper, we present a new network architecture called DRANet which disentangles individual feature representations into two factors, content and style, and transfers domains by applying the style features of another domain.
In contrast to conventional methods which focus on the associations of features among domains, we learn the distinctive features of each domain, then separate the features into two components. 
This design enables us to transfer the domains multi-directionally with our single model.
In addition, our method does not require any class labels for adapting domains.
Another contribution of this work is to propose the a content-adaptive domain transfer method to synthesize more realistic images from the complex scene structures.
Extensive experiments show that our model synthesizes visually pleasing images transferred across domains, and the synthesized images boost the performance of the classification and semantic segmentation tasks.
We also demonstrate that the proposed method outperforms the state-of-the-art domain adaptation methods despite the absence of any labeled data for training.

{\footnotesize
\noindent \textbf{Acknowledgement} This work was supported by the National Research Foundation of Korea (NRF) grant funded by the Korea government (MSIT) (No. 2020R1C1C1013210, No. 2020R1C1C1014863, No. 2018R1A5A1060031), and the DGIST R\&D Program of the Ministry of Science and ICT (20-CoE-IT-01).
}

\clearpage
{\small
\bibliographystyle{ieee_fullname}
\bibliography{egbib}
}

\end{document}